\documentclass{article}
\usepackage{CJKutf8} 
\usepackage{longcat_style}
\usepackage{adjustbox}
\usepackage[utf8]{inputenc} %
\usepackage[T1]{fontenc}    %
\usepackage{newunicodechar}
\usepackage{hyperref}       %
\usepackage{xcolor}
\usepackage[normalem]{ulem} %
\hypersetup{
    colorlinks=true,      %
    linkcolor=blue,      %
    urlcolor=blue,       %
    citecolor=blue,      %
    linkbordercolor=blue, %
    urlbordercolor=blue,
    citebordercolor=blue,
    pdfborderstyle={/S/U/W 1}, %
}

\usepackage{float}
\usepackage{url}            %
\usepackage{booktabs}       %
\usepackage{amsfonts}       %
\usepackage{nicefrac}       %
\usepackage{microtype}      %
\usepackage{lipsum}		%
\usepackage{graphicx}
\usepackage{natbib}
\usepackage{doi}
\usepackage{amsmath}
\usepackage{amssymb} %
\usepackage{xspace}
\usepackage{enumitem}
\usepackage{multirow}
\usepackage{subcaption} 
\usepackage{makecell}
\usepackage{cleveref}
\usepackage{pifont}
\usepackage[inkscapelatex=false]{svg}

\setlist[itemize]{leftmargin=*}
\setlist[enumerate]{leftmargin=*}
\setlist[description]{leftmargin=*}

\newcommand{\tablestyle}[2]{\setlength{\tabcolsep}{#1}\renewcommand{\arraystretch}{#2}\centering\footnotesize}

\usepackage{fancyvrb}
\DefineVerbatimEnvironment{PromptBox}{Verbatim}{frame=single, framesep=2mm, fontsize=\small, commandchars=\\\{\}}

\definecolor{midnightgreen}{rgb}{0.0, 0.29, 0.33}

\title{LARY: A Latent Action Representation Yielding Benchmark for Generalizable Vision-to-Action Alignment}

\author{
  Dujun Nie \quad
  Fengjiao Chen\thanks{Project Lead.} \quad
  Qi Lv \quad 
  Jun Kuang \quad
  Xiaoyu Li \quad 
  Xuezhi Cao \quad 
  Xunliang Cai \\
  Meituan, Beijing, China \\
}

\clearpage

\renewcommand{\headeright}{\raisebox{-0.2\height}{\includegraphics[height=1.8em]{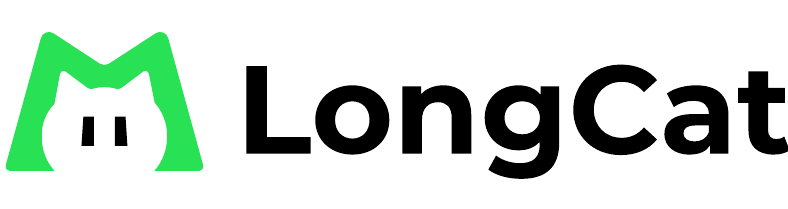}}}
\renewcommand{\shorttitle}{LARY: A Latent Action Representation Yielding Benchmark}

\begin{document}
\maketitle

\begin{abstract}

While the shortage of explicit action data limits Vision-Language-Action (VLA) models, human action videos offer a scalable yet unlabeled data source. A critical challenge in utilizing large-scale human video datasets lies in transforming visual signals into ontology-independent representations, known as latent actions. However, the capacity of latent action representation to derive robust control from visual observations has yet to be rigorously evaluated.
We introduce the Latent Action Representation Yielding (LARY) Benchmark, a unified framework for evaluating latent action representations on both high-level semantic actions (\textit{what to do}) and low-level robotic control (\textit{how to do}). The comprehensively curated dataset encompasses over one million videos (1,000 hours) spanning 151 action categories, alongside 620K image pairs and 595K motion trajectories across diverse embodiments and environments. Our experiments reveal two crucial insights: (i) General visual foundation models, trained without any action supervision, consistently outperform specialized embodied latent action models. (ii) Latent-based visual space is fundamentally better aligned to physical action space than pixel-based space. These results suggest that general visual representations inherently encode action-relevant knowledge for physical control, and that semantic-level abstraction serves as a fundamentally more effective pathway from vision to action than pixel-level reconstruction.

\textbf{GitHub}: \href{https://github.com/meituan-longcat/LARYBench}{https://github.com/meituan-longcat/LARYBench} \\
\textbf{HomePage}: \href{https://meituan-longcat.github.io/LARYBench}{https://meituan-longcat.github.io/LARYBench} \\
\textbf{Hugging Face}: \href{https://huggingface.co/datasets/meituan-longcat/LARYBench}{https://huggingface.co/datasets/meituan-longcat/LARYBench} \\

\end{abstract}

\begin{figure}[h!]
    \centering
    \vspace{-0.9cm}\includegraphics[width=0.8\textwidth]{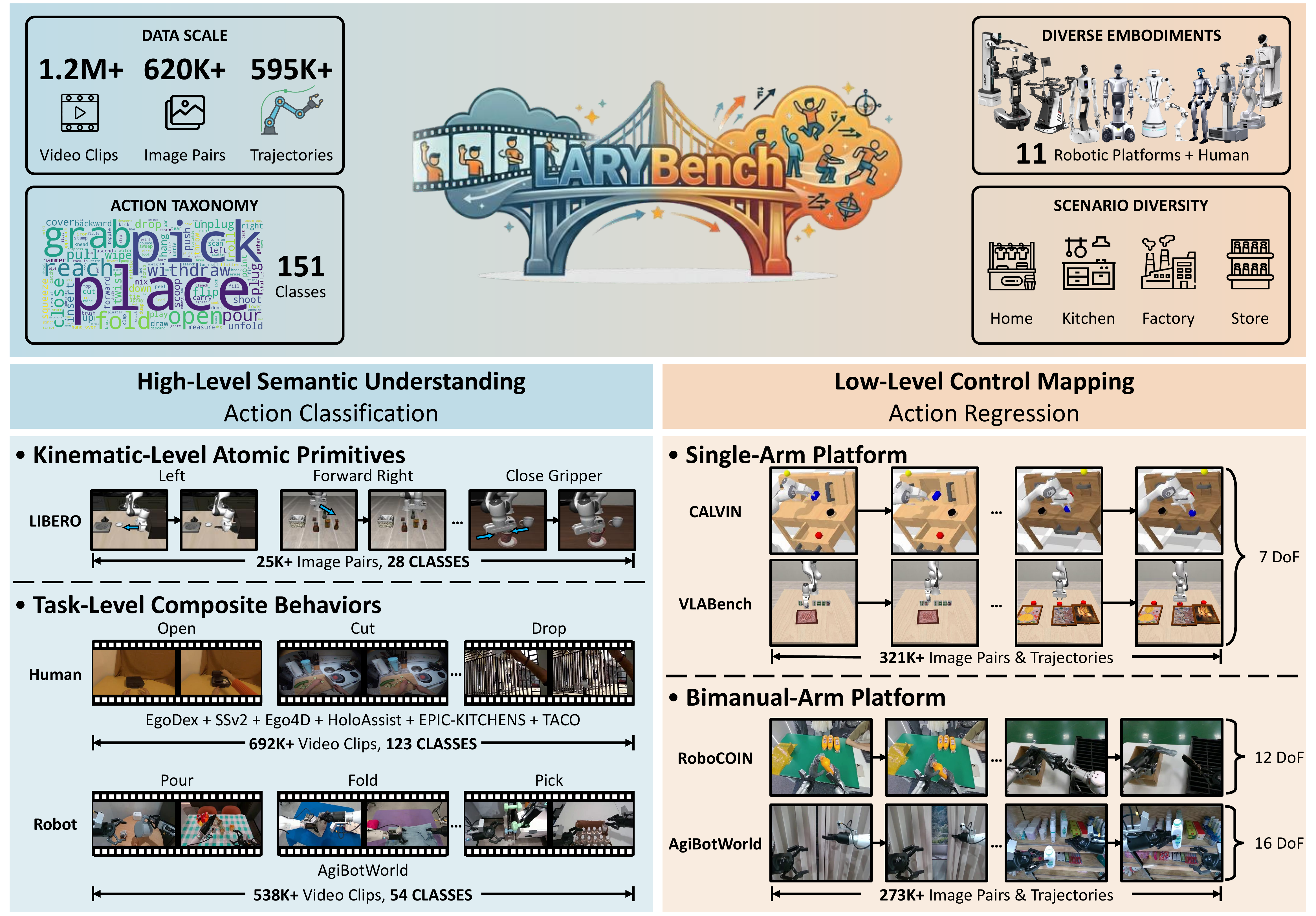}
    \caption{
    LARYBench evaluates vision-to-action transformation on both action generalization and robotic control. 
    }
    \label{fig:lary}
    \vspace{-1.1cm}
\end{figure}

\clearpage
\tableofcontents
\clearpage

\section{Introduction}
The paradigm of learning from large-scale, unlabeled human video data has emerged as a promising solution to the ``data island'' problem in robotics, where diverse action-labeled datasets remain too scarce to train generalist foundation models. 
The pivotal challenge in leveraging human video data lies in the transformation of raw visual signals into ontology-agnostic action representations, commonly referred to as latent actions~\citep{lapa, univla, villa-x, igor, como, moto, gr00t,dreamdojo}.
Pioneering Latent Action Models (LAM) such as Latent Action Pretraining from Videos (LAPA)~\citep{lapa}, Moto~\citep{moto}, and LAPO~\citep{lapo} introduced unsupervised frameworks that tokenize visual changes between video frames into discrete latent action tokens, analogous to word pieces in natural language processing. Similarly, IGOR~\citep{igor} treats the compression of visual changes between initial and goal images as atomic control units, facilitating semantic consistency across human and robot embodiments.

Despite recent progress in transforming vision to latent action, it lacks a thorough and rigorous framework for assessing the quality and effectiveness of latent action representations.
Existing evaluations primarily rely on downstream manipulation task performance or qualitative methods such as cluster visualization~\citep{lapa, univla,adaworld,lapo,igor,moto,videoworld,gr00t}. These approaches fail to decouple the assessment of the VLA components from the latent action quality itself. Furthermore, there is a notable absence of evaluation methods that span different entities, tasks, and granularities, making it difficult to assess the generalization capabilities of action representations. Crucially, the impact of different training architectures, strategies, and usage paradigms on action representation remains underexplored.

To bridge this gap, we introduce the \textbf{L}atent \textbf{A}ction \textbf{R}epresentation \textbf{Y}ielding (\textbf{LARY}) Benchmark, a quantitative framework designed to rigorously evaluate latent action representations. Our objective is to establish a standardized metric that assesses both embodied capabilities spanning cross-agent and cross-scenario applications and video understanding ability. 

Actions in embodied intelligence inherently span two complementary levels: high-level semantic intent that specifies \textit{what to do}, and low-level physical control that determines \textit{how to do it}. 
LARYBench evaluates latent action representations along both dimensions.
For High-Level Semantic Understanding, we assess whether representations can distinguish atomic primitives (e.g., ``move up'', ``close gripper'') and composite behaviors (e.g., ``pick up'', ``place'', ``twist''), drawing on diverse human and robot manipulation videos that cover a wide range of scenarios. 
For Low-Level Control Mapping, we examine whether representations preserve sufficient physical detail to reconstruct end-effector trajectories, using embodied datasets with fine-grained robot motion annotations.

Since current embodied video datasets~\citep{HoloAssist2023, ego4dv2, ssv2, taco, epickitchens, egodex} often suffer from imprecise temporal boundaries and inconsistent action annotations, we develop an automated data engine to re-segment and re-annotate a large-scale corpus. The resulting benchmark comprises over 1.2M short videos (totaling more than 1,000 hours) spanning 151 unique action categories, alongside 620K image pairs and 595K motion trajectories. It covers both human and robotic agents, captured from egocentric and exocentric perspectives across simulated and real-world environments. Building on this diverse foundation, LARYBench instantiates the two evaluation dimensions as concrete tasks: Semantic Action Classification and Low-level Control Regression.


To systematically assess latent action quality, we benchmark three families of models: (i) \textbf{Embodied LAMs} (e.g., LAPA~\citep{lapa}, UniVLA~\citep{univla}, villa-X~\citep{villa-x}), which are purpose-built for robot manipulation; (ii) \textbf{General Vision Encoders}, including both semantic-level (DINOv3~\citep{dinov3}, V-JEPA 2~\citep{vjepa2}) and pixel-level (FLUX.2-dev~\citep{flux2}, Wan2.2~\citep{wan2_2}) backbones, evaluated for their inherent capacity to encode action-relevant features without explicit action supervision; and (iii) \textbf{General LAMs}, a new class of models we propose by grafting the LAM training paradigm onto frozen general vision backbones (e.g., LAPA-DINOv2, LAPA-DINOv3, LAPA-SigLIP2, LAPA-MAGVIT2). Across 11 models, our evaluation reveals a consistent hierarchy: off-the-shelf General Vision Encoders outperform General LAMs, which in turn surpass Embodied LAMs, suggesting that current embodied-specific training not only fails to leverage powerful visual priors but may actually constrain representation quality.

Our contributions are as follows:
    \begin{itemize}
    \item We introduce LARYBench, a comprehensive benchmark that first decouples the evaluation of latent action representations from downstream policy performance. LARYBench probes representations along two complementary dimensions, high-level semantic action (\textit{what to do}) encoding and the low-level physical dynamics required for robotic control (\textit{how to do it}), enabling direct, standardized measurement of representation quality itself.
    \item To support rigorous evaluation, we develop an automated data engine to re-segment and re-annotate a large-scale corpus, yielding 1.2M videos, 620K image pairs, and 595K trajectories across 151 action categories and 11 robotic embodiments, covering both human and robotic agents from egocentric and exocentric perspectives in simulated and real-world environments.
    \item Through systematic evaluation of 11 models, we reveal two consistent findings: (i) action-relevant features can emerge from large-scale visual pre-training without explicit action supervision, and (ii) latent-based feature spaces tend to align with robotic control better than pixel-based ones. These results suggest that future VLA systems may benefit more from leveraging general visual representations than from learning action spaces solely on scarce robotic data.
    \end{itemize}


\section{Related Work}

\textbf{Latent Action Representation from Videos.} 
To alleviate reliance on teleoperated data, recent research extracts latent control signals from unlabeled videos via Inverse Dynamics Models (IDMs). Existing approaches diverge into discrete and continuous paradigms. Discrete methods, such as LAPA~\citep{lapa} and Moto~\citep{moto}, employ Vector Quantization (VQ) to facilitate autoregressive behavior cloning, though often at the cost of fine-grained information loss. Conversely, continuous approaches like CoMo~\citep{como} preserve motion fidelity but risk shortcut learning from background cues. To improve physical grounding and mitigate such artifacts, recent works incorporate semantic constraints via language or saliency (UniVLA~\citep{univla}, IGOR~\citep{igor}) and integrate physical priors, like robot trajectories (LatBot~\citep{latbot}).

\textbf{Latent Actions in World Models and VLAs.}
Latent actions function as a unification interface in generalist systems. In Vision-Language-Action (VLA) architectures like GR00T~\citep{gr00t}, they decouple high-frequency control from low-frequency reasoning. Simultaneously, World Foundation Models (WFMs)—including Cosmos~\citep{cosmos_world}, VideoWorld~\citep{videoworld}, AdaWorld~\citep{adaworld}, and V-JEPA 2~\citep{vjepa2}—utilize latent actions to condition future-frame prediction, enabling agents to internalize physical rules from passive observation. Expanding on the data sources for these models, DreamDojo~\citep{dreamdojo} leverages large-scale human videos to construct a generalist robot world model, demonstrating the efficacy of extracting physical dynamics and actionable representations directly from human demonstrations. Advanced frameworks such as villa-X~\citep{villa-x} refine this by jointly modeling action plans and video generation to ensure semantic alignment between intent and execution.

\textbf{Evaluation of Latent Action Representations.} 
Despite the proliferation of LAMs, quantitative evaluation remains challenging. Standard reconstruction metrics often fail to distinguish action dynamics from environmental noise. While benchmarks like EWMBENCH~\citep{ewmbench} and LAWM~\citep{lawm} utilize trajectory consistency or Canonical Correlation Analysis (CCA) for alignment, recent diagnostic studies~\citep{whatdo} reveal that many models struggle with distractor robustness. Our work extends these inquiries by employing attentive probing and regression to rigorously test the semantic separability and embodied ability of latent action representations.

\section{The LARY Benchmark}

To enable a comprehensive assessment of latent action representations, we propose the Latent Action Representation Yielding Benchmark (LARY), which unifies the evaluation of both high-level semantic action encoding and the low-level physical dynamics required for robotic control.
Formally, given a sequence of visual observations $o_{1:T}$, the motion information is extracted by a latent action model (LAM) as the latents $z \in \mathcal{Z}$. LARYBench evaluates the efficacy of $\mathcal{Z}$ through two tasks: $f_{sem}: \mathcal{Z} \rightarrow \mathcal{C}$ for semantic action decoding via classification task, and $f_{dyn}: \mathcal{Z} \rightarrow \mathcal{A}$ for robotic control construction via regression task.

As shown in Figure~\ref{fig:lary}, LARYBench is curated from a massive amount of multi-embodiment datasets and human datasets, encompassing 151 meticulously defined actions (including 28 atomic actions and 145 composite actions), and corresponding 1.2M annotated samples. The dataset covers a large range of human activities from the frequent ``pick'' and ``place'', to the long-tail ``shovel'' (snow) and ``float'' (balloon). 
To ensure morphological diversity, the dataset spans 11 distinct robotic embodiments, ranging from widely used single-arm manipulators such as Franka to complex bimanual and semi-humanoid platforms such as the AgiBot G1, Agilex Cobot, and Realman series, and includes extensive human-ego-centric interaction data. 
To ensure environmental diversity, the dataset captures thousands of unique object manipulations across a broad spectrum of unstructured environments, including simulated tabletops, authentic residential kitchens, commercial spaces, and industrial scenes.

\begin{figure*}[t]
    \centering 
    \includegraphics[width=\textwidth]{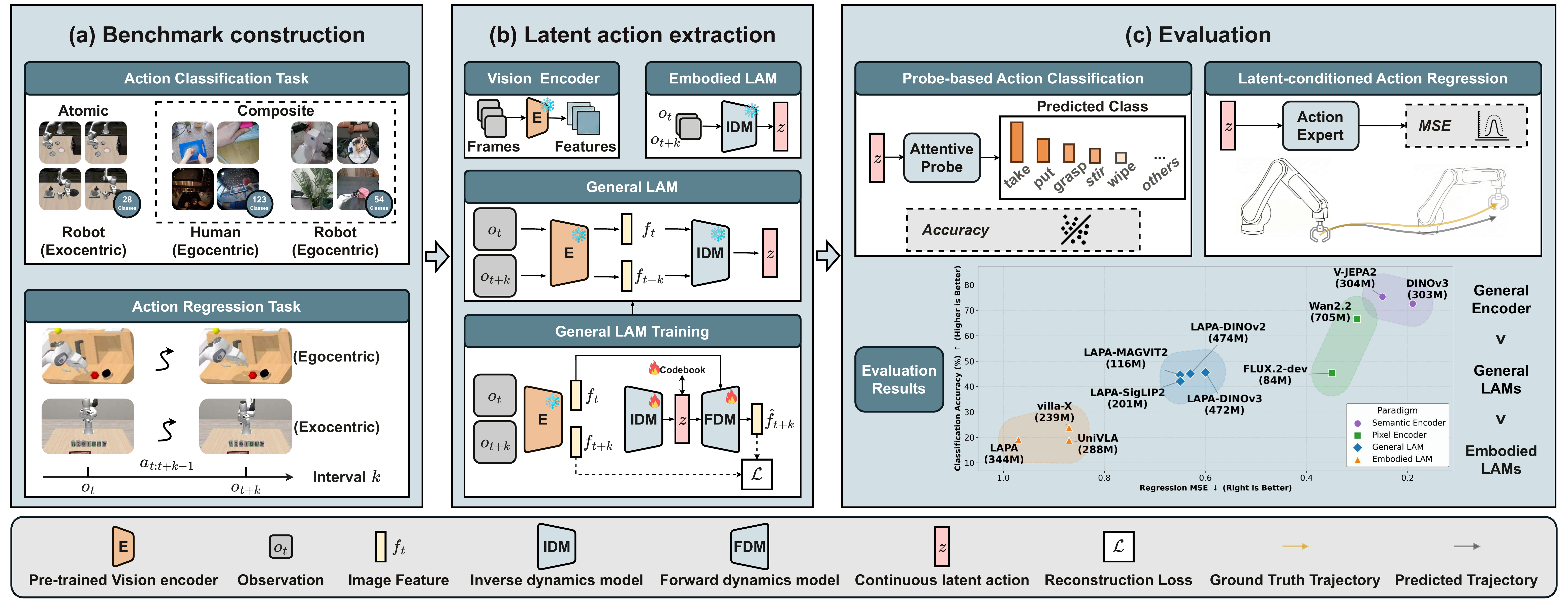}
    \caption{\textbf{Overall pipeline of our benchmark LARY.} First, (a) we construct a comprehensive dataset featuring both atomic and composite actions across varied domains to support classification and regression tasks. Next, (b) we extract the continuous latent action $z$ using various representation paradigms, also highlighting the integration of pre-trained general vision encoders within the VQ-VAE training architecture. Finally, (c) we evaluate these representations utilizing probe-based classification and latent-conditioned regression to quantify their semantic separability (Accuracy) and physical dynamics modeling capabilities (MSE), respectively.}    
    \label{fig:overview}
    \vspace{-0.5cm}

\end{figure*}

With the diverse dataset spanning actions, embodiments, and objects, the evaluation framework of LARYBench is depicted in Figure~\ref{fig:overview}. Next, we introduce the construction pipeline for the semantic action classification task and the robotic control regression task, respectively.

\subsection{Hierarchical Semantic Probing Protocol}
Evaluating the semantic richness of latent action representations requires disentangling spatio-temporal complexities. We formalize this through a multi-granularity semantic probing protocol.

\paragraph{Kinematic-Level Atomic Primitives} At the most granular semantic level, representations must capture instantaneous state variations. We define the \textit{Atomic Robot} task by decomposing the robot’s end-effector actions into 28 discrete kinematic primitives, comprising finely resolved directional translations and binary gripper states. Leveraging exo-centric demonstrations from the LIBERO suite~\citep{libero}, we apply detailed data preprocessing, such as thresholding motion along the $z$-axis relative to static orthogonal directions, to extract 25,940 high-quality image pairs with trajectories.


\paragraph{Task-Level Composite Behaviors} To assess the capacity for abstracting complex behavioral semantics across diverse embodiments and scenarios, we introduce the \textit{Composite Human} and \textit{Composite Robot} tasks. 
To handle diverse data quality across existing datasets~\citep{ssv2,ego4dv2,HoloAssist2023,epickitchens,egodex}, we develop an automated and scalable data engine to perform precise temporal segmentation and semantic alignment (see Figure~\ref{fig:data_pipeline}). Comprehensive details of this engine's architecture are provided in the Appendix. 
By deploying this system across a diverse corpus of ego-centric human datasets (HoloAssist~\citep{HoloAssist2023}, Ego4D~\citep{ego4dv2}, Something-Somethingv2~\citep{ssv2}, TACO~\citep{taco}, EPIC-KITCHENS~\citep{epickitchens}, EgoDex~\citep{egodex}) and realistic bimanual robot demonstrations (AgiBotWorld-Beta~\citep{agibotworld}), we extract 692,297 human clips and 538,423 robot clips under a unified taxonomy of 145 composite behavior classes. Beyond this initial curation, our automated engine holds the potential to continuously process future data streams, ensuring the ever-growing nature of the dataset.
\begin{figure*}[t]
    \center  
    \includegraphics[width=\textwidth]{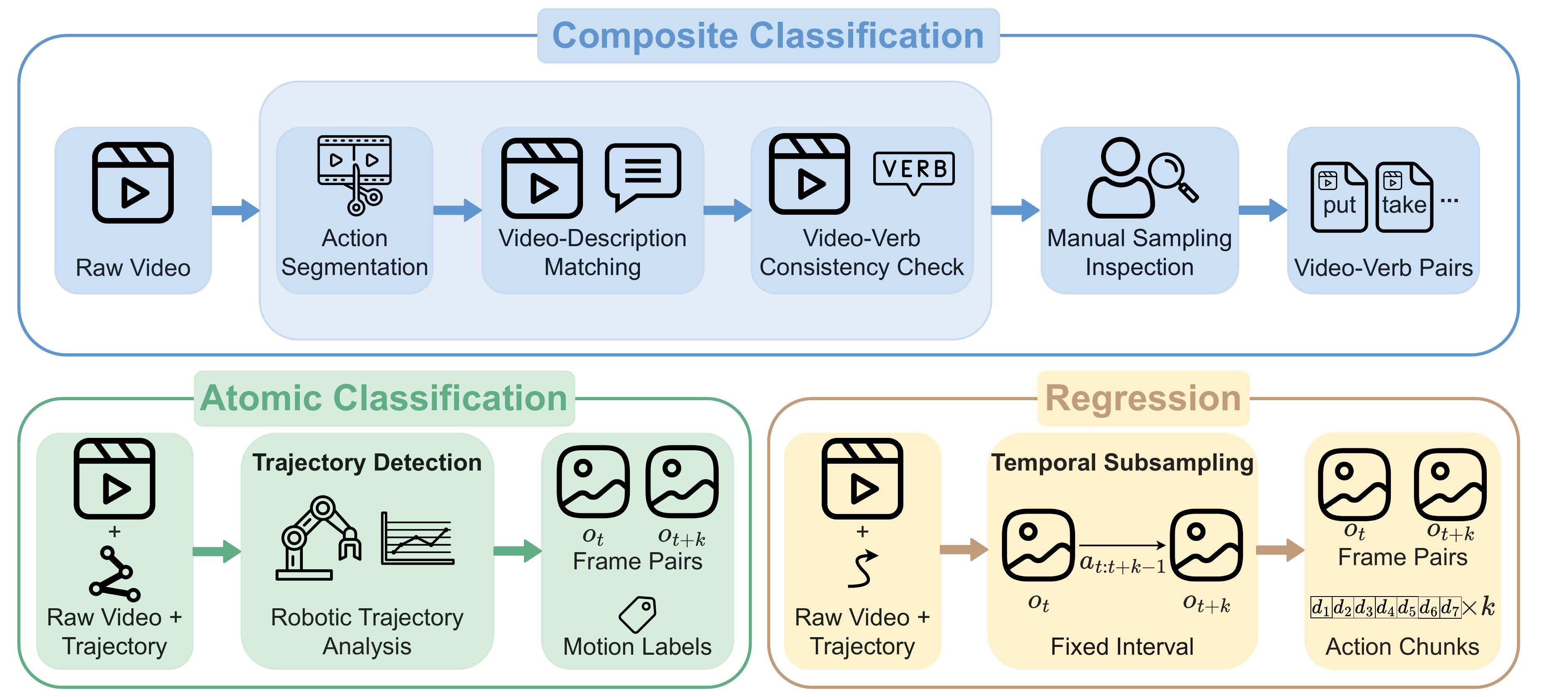}
    \caption{\textbf{Data curation process for LARY-Bench.} To efficiently transform diverse raw videos into standardized evaluation tasks, we integrate a Vision-Language Model (VLM) with robust spatio-temporal video understanding capabilities. The VLM serves as the core reasoning agent for temporal video segmentation and semantic action alignment.}
    \label{fig:data_pipeline}
    \vspace{-0.3cm}
\end{figure*}

\subsection{Physical Execution Mapping Assessment}
While semantic features encode \textit{what to do}, robotic manipulation ultimately demands \textit{how to do}. Our trajectory regression task evaluates the latent space's physical grounding by directly decoding continuous end-effector actions.

This protocol spans diverse hardware morphologies and action spaces. For single-arm exo-centric scenarios, we utilize CALVIN~\citep{calvin} and VLABench~\citep{vlabench} featuring Franka arms. For bimanual ego-centric scenarios, we incorporate RoboCOIN~\citep{robocoin} across 10 diverse platforms and AgiBotWorld-Beta featuring the AgiBot G1. The action space heterogeneity is meticulously preserved: AgiBotWorld-Beta targets a 16-DoF space containing absolute position, quaternions, and gripper state, whereas RoboCOIN maps to a 12-DoF space of position and Euler angles. We deliberately mask the dexterous hand joint data in RoboCOIN to focus the evaluation on macroscopic arm displacements, as fine-grained finger articulation remains an ill-posed inverse problem for current visual encoders.

\section{Experiments}
In this section, we conduct comprehensive experiments and try to answer following questions.
\begin{enumerate}
\item Do Latent Actions Capture Diverse Actions?
\item Do Latent Actions Encode Enough for Control?
\item What Constitutes an Effective Latent Action Model?
\end{enumerate}

\subsection{Taxonomy of Action Representation Paradigms}

To establish a comprehensive baseline, we categorize latent action models into four distinct learning paradigms:

\begin{itemize}
    \item \textbf{Embodied Latent Action Models:} Architectures natively designed for VLAs, such as LAPA~\citep{lapa}, UniVLA~\citep{univla}, and villa-X~\citep{villa-x}, which explicitly entangle temporal dynamics with robotic control via forward/inverse dynamics modeling.
    \item \textbf{General Semantic Encoders:} High-capacity visual backbones like DINOv3~\citep{dinov3} and V-JEPA-2~\citep{vjepa2} optimized for contrastive alignment or latent-level reconstruction, evaluated here for their zero-shot temporal modeling capacity.
    \item \textbf{Generative Pixel Encoders:} Video synthesis models such as Wan2.2 VAE~\citep{wan2_2} and FLUX.2-dev VAE~\citep{flux2} that leverage spatial-temporal compression and pixel-level reconstruction to implicitly capture motion priors.
    \item \textbf{General Latent Action Models:} To leverage the extensive knowledge embedded in pre-trained visual backbones, we explore modeling visual motion at the feature level. Specifically, we substitute the default encoder in the LAPA framework with various pre-trained backbones—ranging from semantic encoders like DINOv2~\citep{dinov2}, DINOv3~\citep{dinov3}, and SigLIP 2~\citep{siglip2} to pixel ones such as MAGVIT2~\citep{openmagvit}. These hybrid models, trained on internet data containing human motions, robot motions, and environment motions, are included to assess how pre-trained visual priors enhance latent action learning.
\end{itemize}

\subsection{Evaluation Settings}

\paragraph{Data Split and Evaluation Metrics}
To maintain consistent class distributions across classification tasks, we apply randomized stratified sampling, partitioning atomic and composite tasks at 75:25 and 70:30 ratios, respectively. Regression tasks on VLABench~\citep{vlabench} adhere to a standard 75:25 train-validation split. For cross-environment evaluation on CALVIN~\citep{calvin}, environments A, B, and C are allocated for training, with D designated for validation. RoboCOIN~\citep{robocoin} and AgiBotWorld-Beta~\citep{agibotworld} are also divided at a 75:25 ratio for training and validation. For semantic action classification, we report the Top-1 Accuracy averaged across all action categories. For low-level control regression, we utilize the Mean Squared Error (MSE) to measure physical fidelity. 

\paragraph{Sampling and Latent Action Extraction}
For the Atomic Robot and regression tasks, we mainly rely on curated image pairs, which allow us to obtain latent action representations from the LAM directly. To retain as much information as possible, we work with the continuous latent embeddings instead of discretized codebook indices, thereby avoiding quantization-related information loss. 
In contrast, extracting representations from video clips in the composite classification tasks is more difficult because the source datasets differ in FPS and exhibit non-uniform motion speeds. Simple uniform sampling generally fails to capture the true temporal dynamics and often yields latent actions that are effectively insensitive to the underlying motion. 
To address this, we employ the Motion-Guided Sampler (MGSampler)~\citep{mgsampler} to select an effective sequence of frames that exhibits sufficient dynamic variation. Latent action features are subsequently extracted across all adjacent sampled frames to ensure the representation effectively encapsulates the transition dynamics.

\paragraph{Probing Protocol for Semantic Action Classification} We uniformly sample 9 frames from each video clip (typically lasting under 5 seconds) and resize every frame to $224 \times 224$. To assess the semantic expressiveness of the learned latent actions, we employ a probe-based classification task. Following the architecture in V-JEPA~\citep{vjepa2}, we use a 4-layer attentive probe as the classifier. Given the varying latent dimensions between general vision encoders and LAMs, we incorporate a projector to align the continuous latent actions into a uniform dimensional space prior to classification, thereby ensuring a fair comparison. The probe is trained for 20 epochs using bfloat16 precision. We apply a multi-head optimization strategy with learning rates from $3 \times 10^{-4}$ to $5 \times 10^{-3}$ and weight decay values from $0.01$ to $0.8$, to ensure a robust evaluation across different representation scales.

\paragraph{Action Experts for Low-level Control Regression} The latent actions are derived from image pairs separated by a frame interval of $s=5$. To assess the physical grounding of latent actions, we implement an easy MLP-based Action Experts to regress absolute end-effector trajectories. It adopts a standard MLP with residual connections architecture featuring 2 residual blocks and a hidden dimension of 4096. This model directly maps latent actions to an action chunk of size $s$ (7-DoF/12-DoF/16-DoF per chunk). 

\paragraph{Feature-Level Latent Action Model Training} To leverage the pretrained vision encoders, we develop a family of Feature-Level Latent Action Models by substituting the visual representation from pixels to features generated by vision encoders in the LAPA framework, while retaining its VQ-VAE structure. During training, the encoder weights remain frozen, and the latent action representation is learned from scratch on an internal video dataset. Detailed training configurations are provided in the Appendix.

\subsection{Benchmark Results}


\paragraph{Do Latent Actions Capture Diverse Actions?}
As shown in Table~\ref{tab:classification}, general vision foundation models (e.g., V-JEPA 2~\citep{vjepa2} and DINOv3~\citep{dinov3}) deliver surprisingly strong performance, even though they do not appear to perform any explicit motion extraction.
\textbf{This phenomenon demonstrates that the visual self-supervised training with large-scale data inherently yields general semantic action representations covering both robot and human data.}
In particular, V-JEPA 2~\citep{vjepa2} learns directly from visual latent features instead of pixel features (the representations commonly used in world models), and achieves the best performance with a substantial margin.
\textbf{This supports the hypothesis that actions can be derived from latent features without needing to be explicitly represented in the pixel space.}
In contrast, existing Embodied LAMs (e.g., LAPA~\citep{lapa}, UniVLA~\citep{univla} and villa-X~\citep{villa-x}) exhibit restricted generalization on diverse actions (averaging 17.99\%–20.90\%), due to the limited amount of training data or the early constraints to low-level actions~\citep{univla, villa-x}.

We try to combine these two kinds of techniques, which extract motion using the self-supervised method in LAPA~\citep{lapa} on general vision foundation models, named as \textbf{General LAMs}. The training dataset is curated from the internet data, containing both human motions and non-human motions (e.g., car driving), which is very different from the test dataset. As shown in Table~\ref{tab:classification}, the General LAMs perform better than existing Embodied LAMs, even on capturing the robot action. Although sharing the same vision foundation model, LAPA-DINOv2 significantly outperforms UniVLA (43.67\% vs. 17.99\%) by leveraging more diverse training data and imposing fewer constraints.
Although the General LAMs achieve relatively high accuracy on human actions, their performance declines on robot actions due to limited data scale and diversity. We leave addressing this issue to future work.

In conclusion, vision foundation models can capture semantic action with self-supervised learning from large-scale of internet visual data. Learning based on latent features performances better than pixel features. More data diversity and fewer task-specific constrains may be the key to generalization.


\paragraph{Do Latent Actions Encode Enough for Control?}
As to the prediction of robotic control, the action regression results show similar performances among candidate models in Table~\ref{tab:regression}, where general vision foundation models achieve significantly better performance among 4 kinds of datasets. 
Especially, latent-based vision encoders (DINOv3~\citep{dinov2} and V-JEPA 2~\citep{vjepa2}) capture better robotic control than pixel-based vision encoders (Wan2.2~\citep{wan2_2} and FLUX.2-dev~\citep{flux2}). 
In other words, \textbf{latent-based visual space is better aligned to robotic action space. }
Considering the good performance of the recent video world model based VLA, which generates pixels first and then decode to robotic control, we suggest the better and more agile way for general robotic control comes from learning latent features. 
We compared DINOv3 with V-JEPA 2~\citep{vjepa2} and found that DINOv3~\citep{dinov3}, thanks to its roots in visual contrastive learning, retains fine-grained recognition capabilities, which in turn improves the precision of fine-grained regression in robotic control tasks.
\begin{table*}[t]
    \tablestyle{4pt}{1.1} 
    \centering
    \small
    \resizebox{\textwidth}{!} 
    {
        \begin{tabular}{@{}l c c cccc@{}} 
            \toprule
            \multirow{2}{*}{\textbf{Model}} & \multirow{2}{*}{\textbf{Paradigm}} & \multirow{2}{*}{\textbf{Params(M)}} & \multicolumn{4}{c}{\textbf{Classification Accuracy$\uparrow$}} \\
            \cmidrule(lr){4-7}
            & & & \textbf{Avg.} & \textit{Atomic Robot} & \textit{Composite Human} & \textit{Composite Robot} \\
            \midrule
            
            V-JEPA2~\citep{vjepa2}          & \multirow{2}{*}{Semantic Encoder}   & 303.89 & \textbf{76.62} & \textbf{79.09} & \textbf{80.35} & \textbf{70.43} \\
            DINOv3~\citep{dinov3}            &                 & 303.13 & \underline{68.68} & \underline{60.79} & \underline{76.19} & \underline{69.06} \\

            \midrule
            Wan2.2~\citep{wan2_2}            & \multirow{2}{*}{Pixel Encoder}                                      & 704.69 & 49.36 & 14.91 & 67.77 & 65.39 \\
            FLUX.2-dev~\citep{flux2}        &                                      & 84.05  & 47.48 & 51.95 & 46.12 & 44.36 \\
            \midrule 
            
            LAPA~\citep{lapa}              & \multirow{3}{*}{Embodied LAM} & 343.80 & 20.17 & 22.27 & 14.61 & 23.64 \\
            UniVLA~\citep{univla}            &                      & 287.75 & 17.99 & 18.62 & 19.08 & 18.56 \\
            villa-X~\citep{villa-x}           &                      & 238.71 & 20.90 & 15.00 & 17.80 & 29.90 \\
            \midrule 

            LAPA-MAGVIT2 &\multirow{4}{*}{General LAM} & 116.40 & 40.78 & 33.12 & 59.70 & 29.53 \\
            LAPA-SigLIP2 &                     & 200.83 & 43.67 & 46.83 & 54.74 & 29.44 \\
            LAPA-DINOv2  &                     & 473.69 & 49.36 & 58.04 & 55.86 & 34.19 \\
            LAPA-DINOv3  &                     & 472.45 & 49.17 & 56.27 & 64.19 & 27.04 \\
            
            \bottomrule
        \end{tabular}
    }
    \caption{\textbf{Action Classification results of latent action representations.} \textbf{Avg.} denotes the mean accuracy across Atomic Robot, Composite Human, and Composite Robot. Best results are in \textbf{bold} and second best are \underline{underlined}.}
    \label{tab:classification}
    \vspace{-10pt}
\end{table*}
\begin{table*}[t]
    \tablestyle{4pt}{1.1} 
    \centering
    \small
    \resizebox{\textwidth}{!} 
    {
        \begin{tabular}{@{}l c c ccccc@{}} 
            \toprule
            \multirow{2}{*}{\textbf{Model}} & \multirow{2}{*}{\textbf{Paradigm}} & \multirow{2}{*}{\textbf{Params(M)}} & \multicolumn{5}{c}{\textbf{Regression MSE$\downarrow$}} \\
            \cmidrule(lr){4-8}
            & & & \textbf{Avg.} & \begin{tabular}{@{}c@{}}CALVIN \end{tabular} & \begin{tabular}{@{}c@{}}VLABench\end{tabular} & \begin{tabular}{@{}c@{}}RoboCOIN\end{tabular} & \begin{tabular}{@{}c@{}}AgiBotWorld-Beta\end{tabular} \\
            \midrule
            V-JEPA 2~\citep{vjepa2}          & \multirow{2}{*}{Semantic Encoder}   & 303.89 & \underline{0.25} & 0.27  & 0.07 & \underline{0.32} & \underline{0.33} \\
            DINOv3~\citep{dinov3}            & & 303.13 & \textbf{0.19} & \textbf{0.22} & \underline{0.06} & \textbf{0.22} & \textbf{0.24} \\
            
            \midrule
            
            Wan2.2~\citep{wan2_2}            & \multirow{2}{*}{Pixel Encoder}                                     & 704.69 & 0.30 & 0.39 & 0.09 & 0.34 & 0.39 \\
            FLUX.2-dev~\citep{flux2}        &                                      & 84.05  & 0.35 & \underline{0.25} & \textbf{0.04} & 0.47 & 0.62 \\
            \midrule 
            
            LAPA~\citep{lapa}              & \multirow{3}{*}{Embodied LAM}        & 343.80 & 0.97 & 0.96 & 0.95 & 0.96 & 1.00 \\
            UniVLA~\citep{univla}            &                                      & 287.75 & 0.87 & 0.82 & 0.74 & 0.94 & 0.97 \\
            villa-X~\citep{villa-x}           &                                      & 238.71 & 0.87 & 0.86 & 0.72 & 0.94 & 0.97 \\
            \midrule 

            LAPA-MAGVIT2 &\multirow{4}{*}{General LAM}         & 116.40 & 0.65 & 0.59 & 0.36 & 0.80 & 0.83 \\
            LAPA-SigLIP2 &                                     & 200.83 & 0.65 & 0.57 & 0.30 & 0.86 & 0.88 \\
            LAPA-DINOv2  &                                     & 473.69 & 0.63 & 0.55 & 0.26 & 0.85 & 0.86 \\
            LAPA-DINOv3  &                                     & 472.45 & 0.60 & 0.50 & 0.25 & 0.82 & 0.84 \\
            
            \bottomrule
        \end{tabular}
    }
    \caption{\textbf{Action Regression results of latent action representations.} \textbf{Avg.} is the mean MSE across all datasets. Best results are in \textbf{bold} and second best are \underline{underlined}.}
    \label{tab:regression}
    \vspace{-10pt}
\end{table*}

\begin{figure*}[t]
    \centering
    \includegraphics[width=\textwidth]{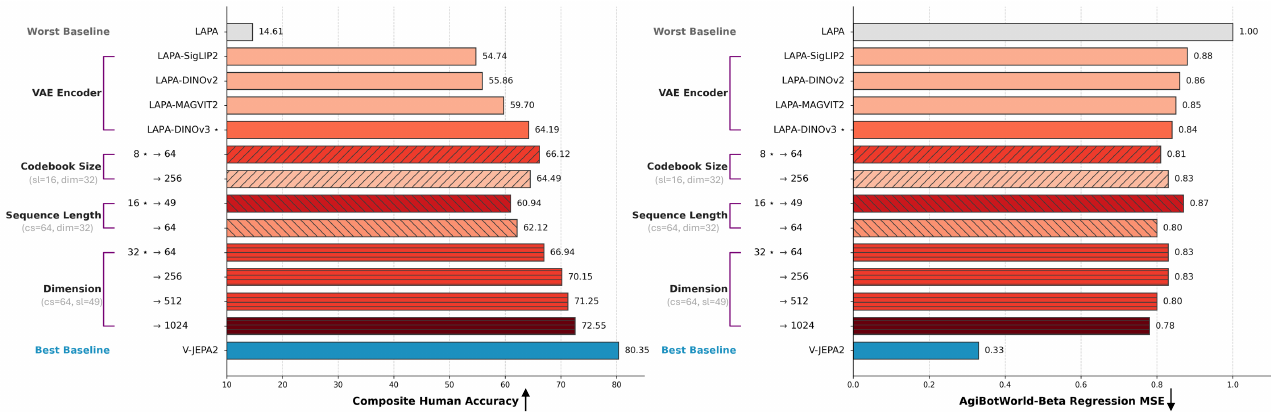}
    \caption{\textbf{Performance Evolution of Latent Action Models.} The (*) denotes the default quantization settings for LAPA-DINOv3 ($cs=8, sl=16, dim=32$). $cs$, $sl$, and $dim$ represent Codebook Size, Sequence Length and Latent Dimension respectively. \textit{Composite Human} classification on the left and AgiBotWorld-Beta~\cite{agibotworld} regression on the right.}
    \label{fig:ablation_lam}
    \vspace{-0.3cm}

\end{figure*}

\paragraph{What Constitutes an Effective Latent Action Model?}
\label{sec:lam_train}

To systematically build a robust latent action model, we conduct ablations under the LAPA~\citep{lapa} framework. Figure~\ref{fig:ablation_lam} maps a performance evolution path bridging the gap between the worst baseline (LAPA) and the continuous upper bound (V-JEPA2). First, self-supervised visual encoders (e.g., DINOv3~\citep{dinov3}) consistently construct superior LAMs compared to reconstruction-based and vision-language contrastive models like MAGVIT2~\citep{openmagvit} and SigLIP2~\citep{siglip2}(Tables~\ref{tab:classification}-\ref{tab:regression}), indicating that contrastive priors better capture fine-grained spatial-temporal correspondences. Setting LAPA-DINOv3 ($cs=8, sl=16, dim=32$) as our foundation, we sequentially optimize its quantization bottleneck. Second, expanding codebook capacity improves downstream regression, but overly large sizes (e.g., 256) cause codebook utilization drops without further gains (Table~\ref{tab:ablation_cs}), making a moderate size ($cs=64$) optimal for dense representations. Third, sequence length critically dictates temporal diversity; short sequences ($sl=16$) trigger catastrophic codebook collapse (1.6\% utilization), whereas a moderate length ($sl=49$) ensures 100\% utilization and robust generalization (Table~\ref{tab:ablation_sl}). Finally, scaling the latent dimension improves theoretical capacity but introduces quantization instability, evidenced by sudden utilization collapses at intermediate sizes (Table~\ref{tab:ablation_dim}). Thus, $dim=256$ strikes the best capacity-stability balance. \textbf{Ultimately, beyond dataset quality, an effective latent action space requires two key components: robust self-supervised visual priors to capture precise spatial-temporal dynamics, and a strictly regularized quantization bottleneck to maintain stability and dense utilization}.

\begin{table*}[t]
    \centering
    \small
    \caption{\textbf{Ablation on Codebook Size.} Default settings are Sequence Length = 49, Latent Dimension = 256. Best results are in \textbf{bold} and second best are \underline{underlined}.}
    \label{tab:ablation_cs}
    \resizebox{0.9\textwidth}{!}{
        \begin{tabular}{@{} l c c c c c @{}}
            \toprule
            \multirow{2}{*}{\makecell{\textbf{Codebook} \\ \textbf{Size}}} & \multirow{2}{*}{\makecell{\textbf{Recon} \\ \textbf{Loss}}} & \multirow{2}{*}{\makecell{\textbf{Codebook} \\ \textbf{Utilization (\%)}}} & \multicolumn{2}{c}{\textbf{Accuracy (Composite task)$\uparrow$}} & \multirow{2}{*}{\makecell{\textbf{MSE} \\ \textbf{(AgiBotWorld-Beta)$\downarrow$}}} \\
            \cmidrule(lr){4-5}
            & & & Human & Robot & \\
            \midrule
            8   & 0.00808 & \textbf{100.0} & \textbf{71.31} & \textbf{64.89} & 0.88 \\
            64  & \textbf{0.00751} & \textbf{100.0} & \underline{70.15} & \underline{64.04} & \textbf{0.83} \\
            256 & \underline{0.00758} & \underline{89.5}  & 69.84 & 63.82 & \underline{0.85} \\
            \bottomrule
        \end{tabular}
    }
\end{table*}

\begin{table*}[t]
    \centering
    \small
    \caption{\textbf{Ablation on Sequence Length.} Default settings are Codebook Size = 64, Latent Dimension = 256. Best results are in \textbf{bold} and second best are \underline{underlined}.}
    \label{tab:ablation_sl}
    \resizebox{0.9\textwidth}{!}{
        \begin{tabular}{@{} l c c c c c @{}}
            \toprule
            \multirow{2}{*}{\makecell{\textbf{Sequence} \\ \textbf{Length}}} & \multirow{2}{*}{\makecell{\textbf{Recon} \\ \textbf{Loss}}} & \multirow{2}{*}{\makecell{\textbf{Codebook} \\ \textbf{Utilization (\%)}}} & \multicolumn{2}{c}{\textbf{Accuracy (Composite task)$\uparrow$}} & \multirow{2}{*}{\makecell{\textbf{MSE} \\ \textbf{(AgiBotWorld-Beta)$\downarrow$}}} \\
            \cmidrule(lr){4-5}
            & & & Human & Robot & \\
            \midrule
            16  & 0.00908 & 1.6   & 69.23 & 63.33 & \underline{0.79} \\
            49  & \textbf{0.00751} & \textbf{100.0} & \underline{70.15} & \underline{64.04} & 0.83 \\
            64  & \underline{0.00773} & \underline{79.7}  & \textbf{71.37} & \textbf{64.70} & \textbf{0.72} \\
            \bottomrule
        \end{tabular}
    }
\end{table*}

\begin{table*}[t]
    \centering
    \small
    \caption{\textbf{Ablation on Latent Dimension.} Default settings are Codebook Size = 64, Sequence Length = 49. Best results are in \textbf{bold} and second best are \underline{underlined}.}
    \label{tab:ablation_dim}
    \resizebox{0.9\textwidth}{!}{
        \begin{tabular}{@{} l c c c c c @{}}
            \toprule
            \multirow{2}{*}{\makecell{\textbf{Latent} \\ \textbf{Dimension}}} & \multirow{2}{*}{\makecell{\textbf{Recon} \\ \textbf{Loss}}} & \multirow{2}{*}{\makecell{\textbf{Codebook} \\ \textbf{Utilization (\%)}}} & \multicolumn{2}{c}{\textbf{Accuracy (Composite task)$\uparrow$}} & \multirow{2}{*}{\makecell{\textbf{MSE} \\ \textbf{(AgiBotWorld-Beta)$\downarrow$}}} \\
            \cmidrule(lr){4-5}
            & & & Human & Robot & \\
            \midrule
            32   & 0.01141 & 3.1   & 60.94 & 57.69 & 0.87 \\
            64   & 0.00967 & \textbf{100.0} & 66.94 & 63.07 & 0.83 \\
            256  & 0.00751 & \textbf{100.0} & 70.15 & 64.04 & 0.83 \\
            512  & \underline{0.00732} & 1.6   & \underline{71.25}  & \underline{64.97}& \textbf{0.80} \\
            1024 & \textbf{0.00670} & \underline{84.4}  & \textbf{72.55} & \textbf{65.78}& \underline{0.81} \\
            \bottomrule
        \end{tabular}
    }
\end{table*}

\section{Error Analysis}
To provide deeper insights into the limitations of latent action representations, we conduct a fine-grained error decomposition.

\subsection{The Long-Tail Dilemma and Mid-Frequency Semantic Aliasing}
\begin{figure}[h!]
    \centering
    \includegraphics[width=\textwidth]{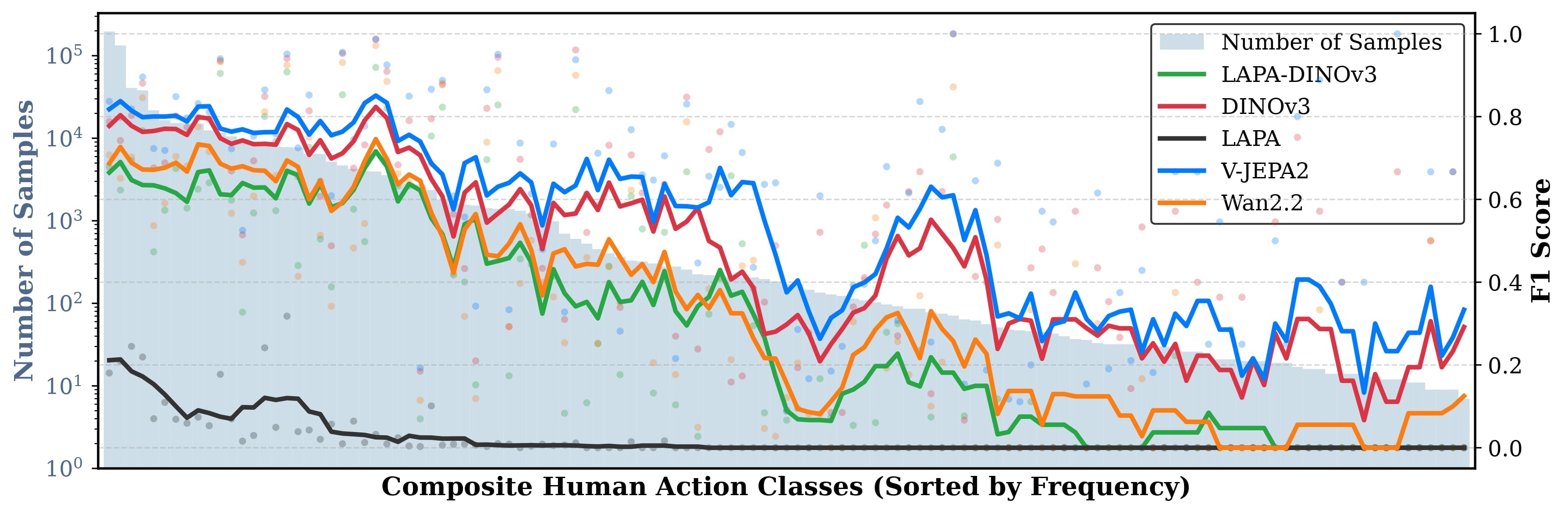}
    \vspace{-0.3cm}
    \caption{\textbf{Action classification performance across the long-tail distribution of the \textit{Composite Human} dataset.} The background histogram indicates the number of samples per action class, sorted by descending frequency. Solid lines represent the moving average F1 scores, while transparent scatters denote exact class-wise scores.}
    \label{fig:long_tail}
    \vspace{-0.5cm}
\end{figure}

Figure~\ref{fig:long_tail} illustrates action classification performance across class frequencies on the \textit{Composite Human} dataset. While the baseline LAPA~\citep{lapa} model exhibits uniformly poor performance, LAPA-DINOv3 closely mirrors the continuous DINOv3~\citep{dinov3} encoder, demonstrating a direct inheritance of both its representational strengths and specific vulnerabilities. 
Stronger models generally outperform weaker ones across most actions, suggesting that the LARYBench provides a stable and reliable evaluation of model capabilities.
As the frequency decreases, the performance gap between strong and weak models widens, indicating that strong models exhibit better generalization capabilities in long-tail scenarios.


\subsection{Spatiotemporal Grounding and Action-Centric Attention}
Figure~\ref{fig:attn_map} visualizes the cross-attention maps (additional heatmap visualizations are provided in the Appendix), revealing a stark contrast in how different models ground physical interactions. First, among general visual encoders, self-supervised models demonstrate superior spatiotemporal grounding. Notably, V-JEPA 2~\citep{vjepa2} exhibits the most accurate attention distribution, sharply localizing on the exact interaction points between both the left and right hands and the manipulated object (bowl). DINOv3~\citep{dinov3} similarly maintains a precise, geometry-aware focus on the active end-effector. 
Conversely, generative priors Flux2-dev~\citep{flux2} and Wan2.2~\citep{wan2_2} exhibit highly dispersed, unfocused attention distributions, indicating an inherent bias toward global scene understanding rather than localized physical interactions. Second, regarding LAMs versus general encoders, standard Embodied LAMs, like LAPA~\citep{lapa}, completely fail to localize meaningful features, producing broad, uninformative attention blobs. However, our LAPA-DINOv2 effectively inherits the strong localization capabilities of its backbone. Despite the severe spatial quantization bottleneck (e.g., $SL=16$, producing coarse $4\times4$ patches), it successfully anchors its attention to the active object. Finally, within the LAM architectures, our General LAMs consistently exhibit sharper object-centric grounding compared to other Embodied LAMs (e.g., UniVLA~\citep{univla}, villa-X~\citep{villa-x}), which suffer from severe attention diffusion. \textbf{Ultimately, this demonstrates that predictive failure fundamentally stems from an inability to concentrate attention on the corresponding dynamically interacting objects}.

\begin{figure}[t]
    \centering
    \vspace{-0.3cm}
    \includegraphics[width=\textwidth]{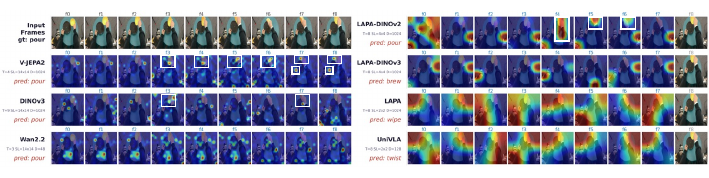}
    \vspace{-0.3cm}
    \caption{\textbf{Cross-attention heatmaps of the temporal pooler across various models for a 9-frame \textit{pour} sequence.} The spatial sequence length ($SL$) dictates visual granularity: $SL=14*14$ yield fine-grained $14\times14$ heatmaps, whereas LAMs ($SL \in \{4, 16\}$) produce blocky $2\times2$ or $4\times4$ attention regions. Temporally, $T=9$ models compute frame-by-frame attention; $T=8$ models extract pairwise latent actions (leaving the 9th frame unmodified); and models with temporal compression ($T \in \{3,4\}$) duplicate attention maps across their corresponding frame groups.}
    \label{fig:attn_map}
    \vspace{-0.5cm}
\end{figure}

\subsection{Stride Ablation: Latent Action Spaces Encode Robust Dynamic Trajectories}
To investigate the temporal robustness of various representations, we ablate the sampling stride (stride=5, 15, 30) between input frames on VLABench, as shown in Table~\ref{tab:stride_ablation}. Crucially, as the stride increases, the number of actions to regress scales linearly, significantly amplifying the dimensionality and complexity of the task. Pure spatial generative encoders, such as FLUX.2-dev~\citep{flux2}, excel at extremely short horizons (achieving 0.04 MSE at stride=5) but fail catastrophically under this increased temporal dimensionality (MSE spiking to 0.62 at stride=30). In contrast, Latent Action Models (LAMs)—encompassing both Embodied LAMs and our General LAMs—exhibit consistent stability across all strides. This shared characteristic proves that the latent action paradigm does not merely encode static spatial alignments; instead, it successfully captures and preserves underlying dynamic action trajectories. While traditional Embodied LAMs suffer from a high error (about 0.70 average MSE), introducing general visual priors as in our General LAMs effectively lowers this error margin while fully inheriting the paradigm's temporal stability. \textbf{In conclusion, although a performance gap remains when compared to the absolute regression accuracy of uncompressed general vision encoders, the fundamental mechanism of mapping visual observations into a latent action space provides a uniquely robust representation for long-horizon intents, proving its inherent necessity for continuous control tasks.}
\begin{table*}[t]
    \tablestyle{4pt}{1.1} 
    \centering
    \small
    \resizebox{0.9\textwidth}{!} 
    {
        \begin{tabular}{l c c cccc} 
            \toprule
            \multirow{2}{*}{\textbf{Model}} & \multirow{2}{*}{\textbf{Paradigm}} & \multirow{2}{*}{\textbf{Params(M)}} & \multicolumn{4}{c}{\textbf{Regression MSE}$\downarrow$} \\
            \cmidrule(lr){4-7} 
            & & & stride=5 & stride=15 & stride=30 & \textbf{Avg.} \\
            \midrule
            
            V-JEPA 2~\citep{vjepa2}          & \multirow{2}{*}{Semantic Encoder}   & 303.89 & 0.07 & \textbf{0.13} & \textbf{0.16} & \textbf{0.12} \\
            DINOv3~\citep{dinov3}            &                 & 303.13 & \underline{0.06} & 0.20 & 0.25 & 0.17 \\
            \midrule
            
            Wan2.2~\citep{wan2_2}            & \multirow{2}{*}{Pixel Encoder}                                     & 704.69 & 0.09 & \underline{0.16} & \underline{0.24} & \underline{0.16} \\
            FLUX.2-dev~\citep{flux2}        &                                      & 84.05  & \textbf{0.04} & 0.57 & 0.62 & 0.41 \\
            \midrule 
            
            LAPA~\citep{lapa}              & \multirow{3}{*}{Embodied LAM} & 343.80 & 0.95 & 0.87 & 0.77 & 0.86 \\
            UniVLA~\citep{univla}            &                      & 287.75 & 0.74 & 0.68 & 0.69 & 0.70 \\
            villa-X~\citep{villa-x}           &                      & 238.71 & 0.72 & 0.70 & 0.66 & 0.69 \\
            \midrule 

            LAPA-MAGVIT2 &\multirow{4}{*}{General LAM} & 116.40 & 0.36 & 0.32 & 0.33 & 0.34 \\
            LAPA-SigLIP2 &                     & 200.83 & 0.30 & 0.25 & \underline{0.24} & 0.26 \\
            LAPA-DINOv2  &                     & 473.69 & 0.26 & 0.23 & 0.37 & 0.29 \\
            LAPA-DINOv3  &                     & 472.45 & 0.25 & 0.20 & 0.26 & 0.24 \\
            
            \bottomrule
        \end{tabular}
    }
    \caption{\textbf{Ablation study on VLABench~\citep{vlabench} sampling stride.} We report the regression MSE with sampling strides of 5, 15, and 30, and their average. Best results are in \textbf{bold} and second best are \underline{underlined}.}
    \label{tab:stride_ablation}
    \vspace{-10pt}
\end{table*}

\section{Conclusion}
We introduce LARYBench to evaluate latent action representations across kinematic and semantic granularities. Our empirical results reveal that general visual foundation models consistently outperform specialized Embodied Latent Action Models (LAMs). Specifically, visual understanding models dominate in semantic tasks, while general visual encoders surprisingly achieve superior regression MSE without domain-specific training. This indicates that effective latent actions naturally emerge from large-scale visual pre-training, whereas specialized Embodied LAMs often suffer from representation collapse due to data scarcity or premature constraints to domain-specific low-level control. Consequently, we advocate for a paradigm shift in Vision-Language-Action (VLA) design: instead of learning action spaces from limited robotic data, future research should focus on aligning control policies with the robust feature spaces of general vision models. Addressing architectural challenges such as continuous signal decoding and feature alignment will be pivotal to unlocking these universal priors for data-efficient embodied agents.

\section*{Acknowledgement}

We hereby express our appreciation to the LongCat Team EVA Committee for their valuable assistance, guidance, and suggestions throughout the course of this work.

\bibliographystyle{unsrtnat}
\bibliography{main}

\clearpage

\appendix

\setcounter{table}{0}
\setcounter{figure}{0}

\section{Overview of the Appendix}

This Appendix is organized as follows:
\begin{itemize}
    \item Section~\ref{sec:LARYBench_Details} contains details about the composition of LARYBench and the automated data curation process;
    \item Section~\ref{sec:Experiment_Details} contains experimental details, including the compute environment, model training configurations, and hyperparameter settings;
    \item Section~\ref{sec:Additional_Visualizations} contains extended cross-attention heatmaps and additional case studies for different tasks.
\end{itemize}

\section{Details of LARYBench}
\label{sec:LARYBench_Details}

\subsection{Composition of LARYBench}
LARYBench comprises over 1.2 million short videos (totaling $>$1,000 hours), 620K image pairs, and 595K motion trajectories. Figure~\ref{fig:data_composition_pie} illustrates the overall proportion of these diverse data sources and duration of the video clips. The benchmark evaluates representations across 151 meticulously curated unique action categories to assess both high-level semantic intent and low-level physical execution. We now provide a detailed breakdown of LARYBench by task track:

\begin{figure}[htbp]
    \centering
    \includegraphics[width=\linewidth]{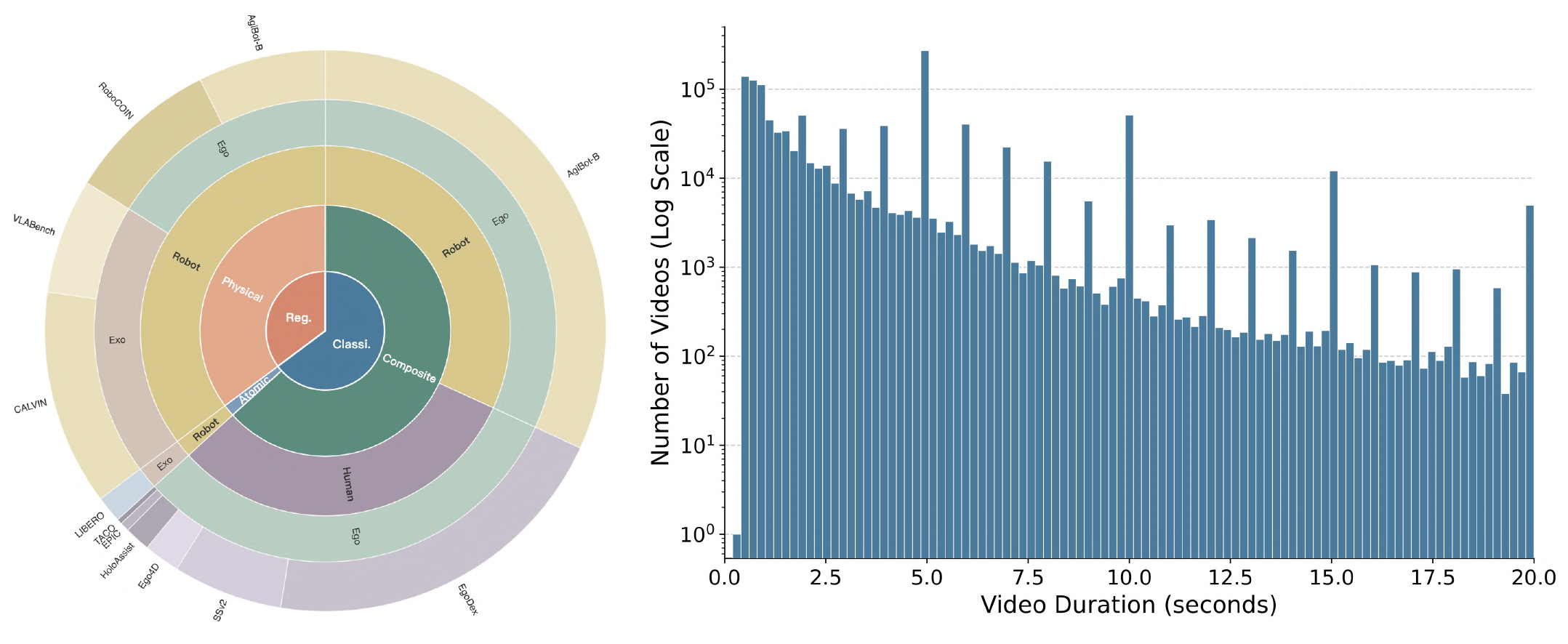}
    \caption{Left: Overall composition of the LARYBench dataset. Reg. and Classi. represent regression and classification. Ego and Exo represent Egocentric and Exocentric. AgiBot-B, EPIC, and SSv2 represent AgiBotWorld-Beta~\citep{adaworld}, EPIC-KITCHENS~\citep{epickitchens}, and Something-Somethingv2~\citep{ssv2}. Right: Duration distribution of all videos (Composite tasks).}
    \label{fig:data_composition_pie}
\end{figure}

\begin{figure}[htbp]
    \centering
    \includegraphics[width=\linewidth]{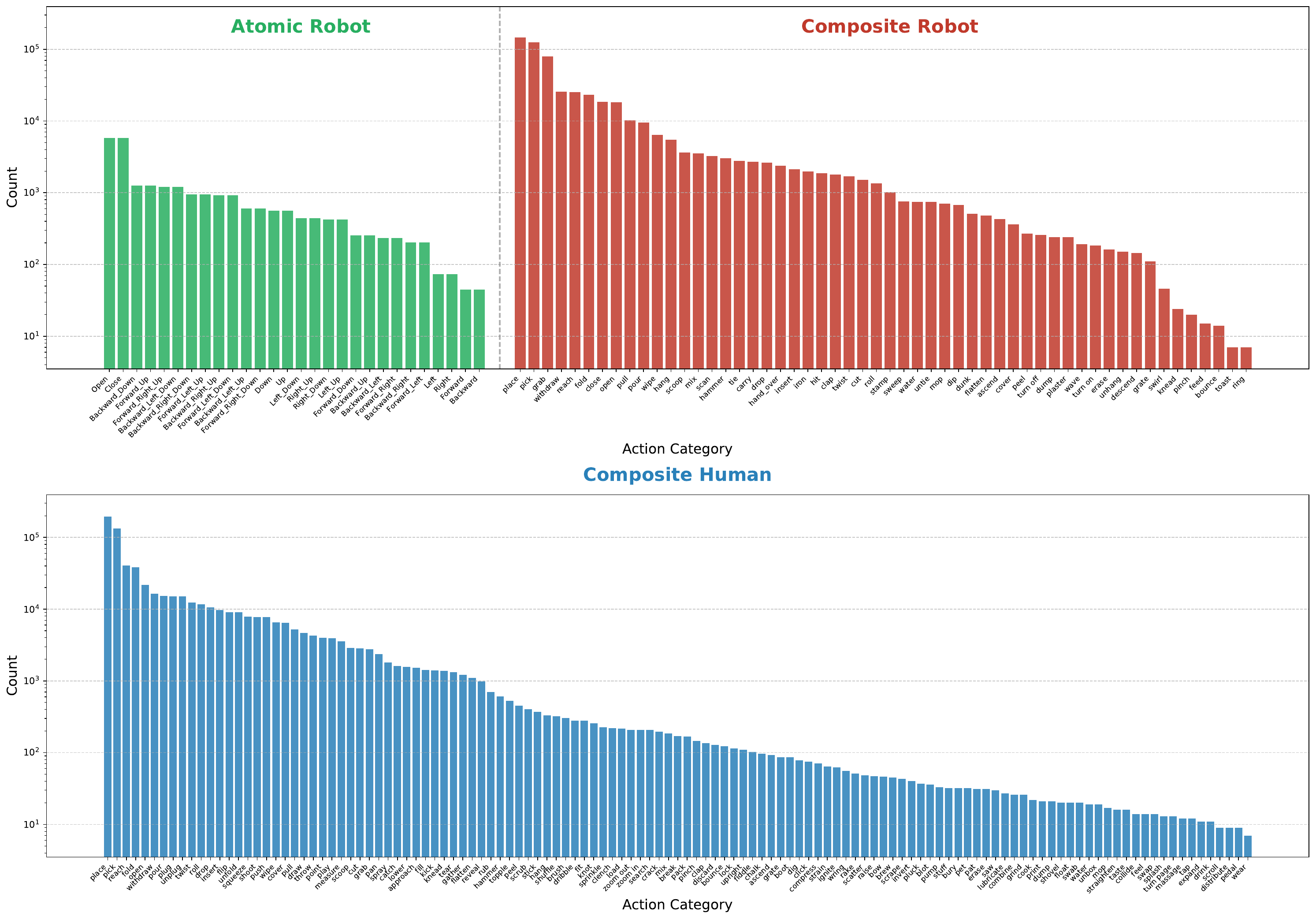}
    \caption{Distribution of the action categories of the \textit{Atomic Robot}, \textit{Composite Human}, and \textit{Composite Robot} datasets.}
    \label{fig:action_distribution}
\end{figure}

\begin{figure}[htbp]
    \centering
    \begin{minipage}{0.5\linewidth}
        \centering
        \includegraphics[width=\linewidth]{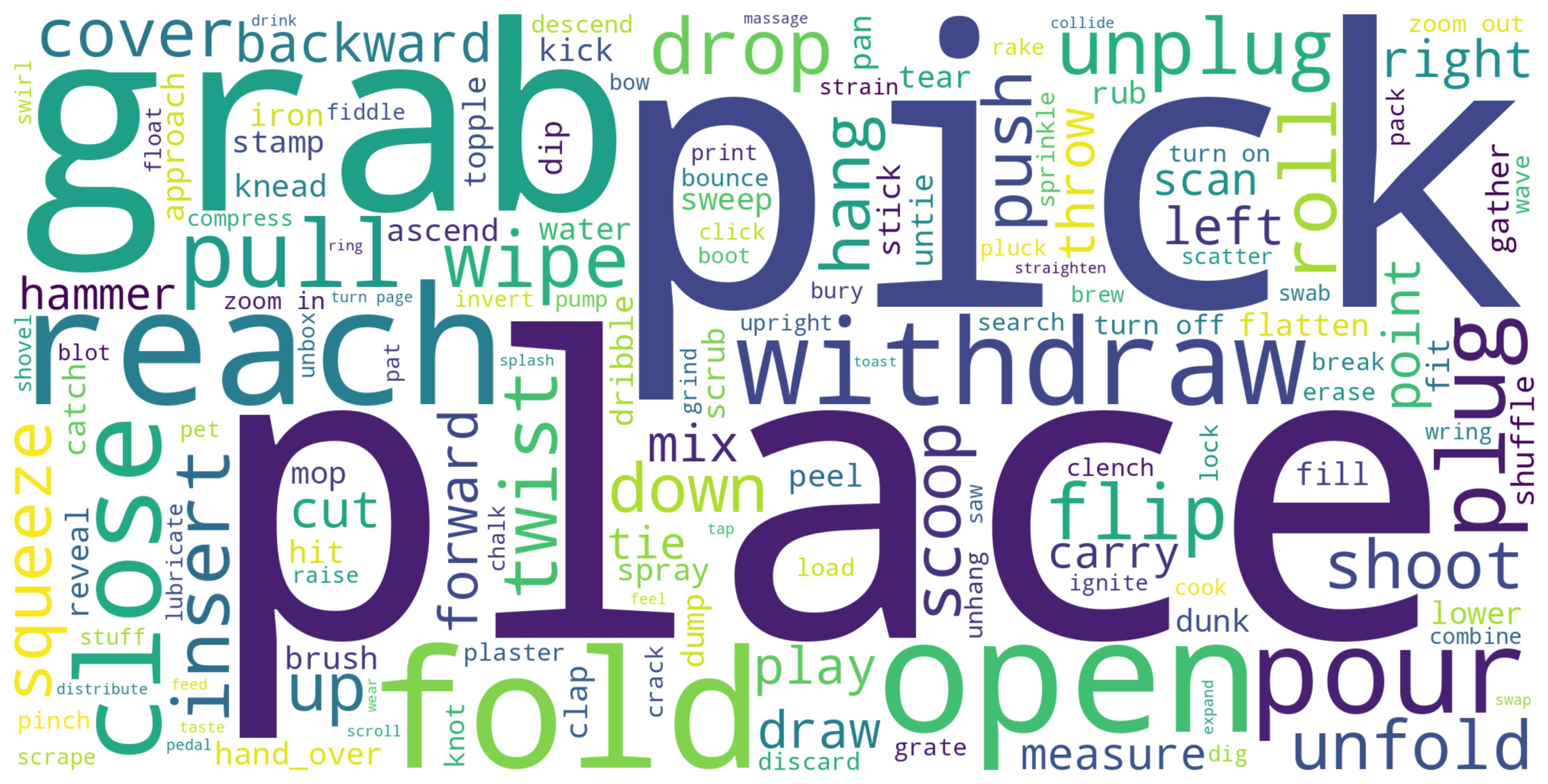}
        \centerline{(a) Verb Word Cloud}
    \end{minipage}\hfill
    \begin{minipage}{0.5\linewidth}
        \centering
        \includegraphics[width=\linewidth]{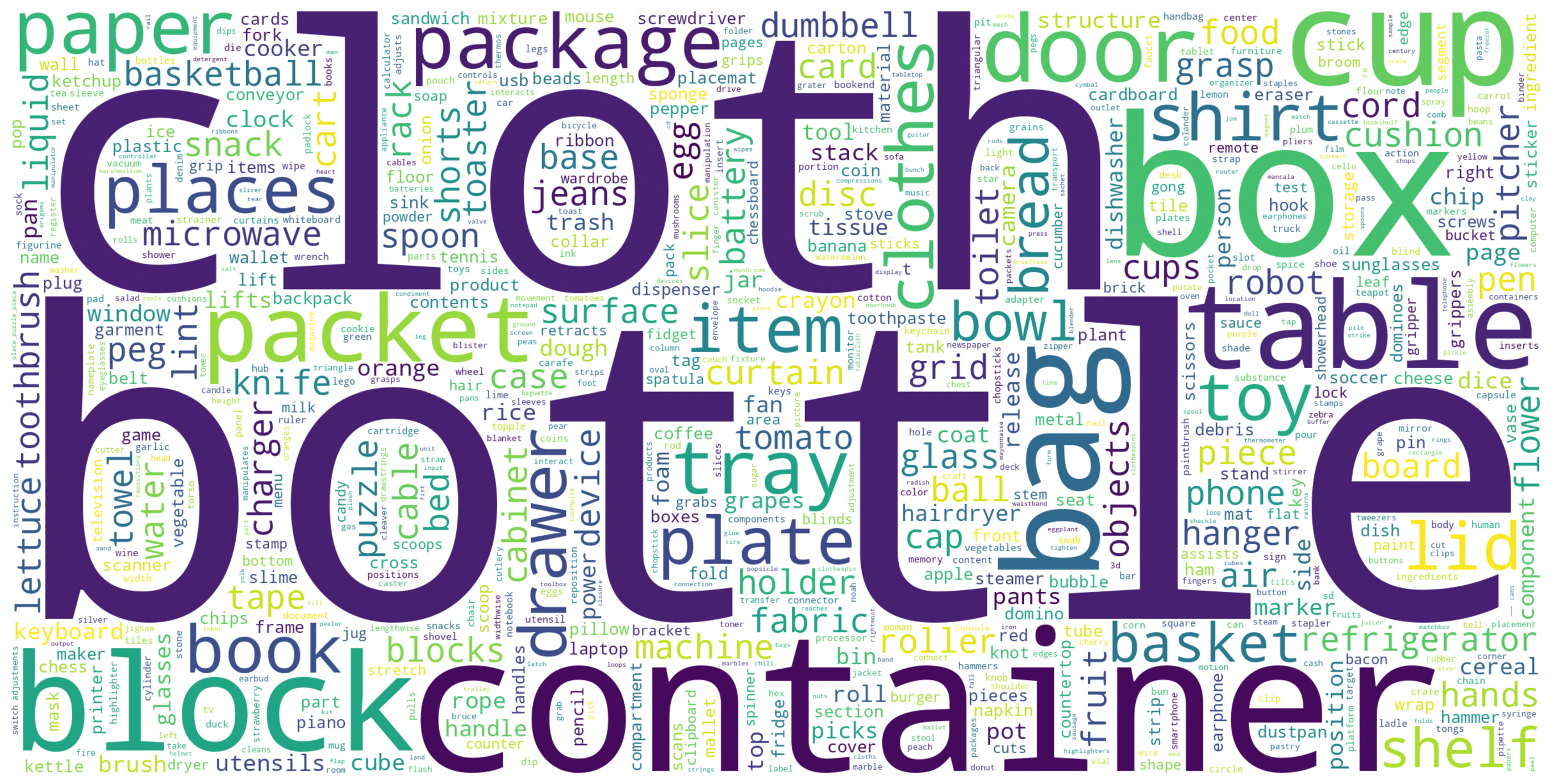}
        \centerline{(b) Noun Word Cloud}
    \end{minipage}
    \caption{Word clouds illustrating the semantic diversity of the LARYBench dataset annotations. (a) Highlights the wide variety of actionable verbs indicating the interaction types, while (b) showcases the extensive range of nouns representing the manipulated objects across both human and robotic domains.}
    \label{fig:word_clouds}
\end{figure}

\paragraph{Classification Track}
This track evaluates the classification capabilities of latent representations, encompassing both abstract behavioral semantics and fine-grained kinematic changes. While the subsets detailed below sum to more than 151 classes individually (28 \textit{Atomic Robot} classes, 54 \textit{Composite Robot} classes, and 123 \textit{Composite Human} classes), substantial semantic overlap exists across embodiments (e.g., both human and robotic domains contain shared actions like \textit{pick} and \textit{place}). Consequently, the unified taxonomy encompasses exactly 151 unique action categories. To illustrate the scale and diversity of the dataset, Figure~\ref{fig:action_distribution} presents the sample frequency across all categories. Furthermore, to highlight the rich semantic landscape of the physical interactions and manipulated objects, Figure~\ref{fig:word_clouds} visualizes the distribution of verbs and nouns across the dataset annotations.

\begin{figure}[htbp]
    \centering
    \includegraphics[width=0.8\linewidth]{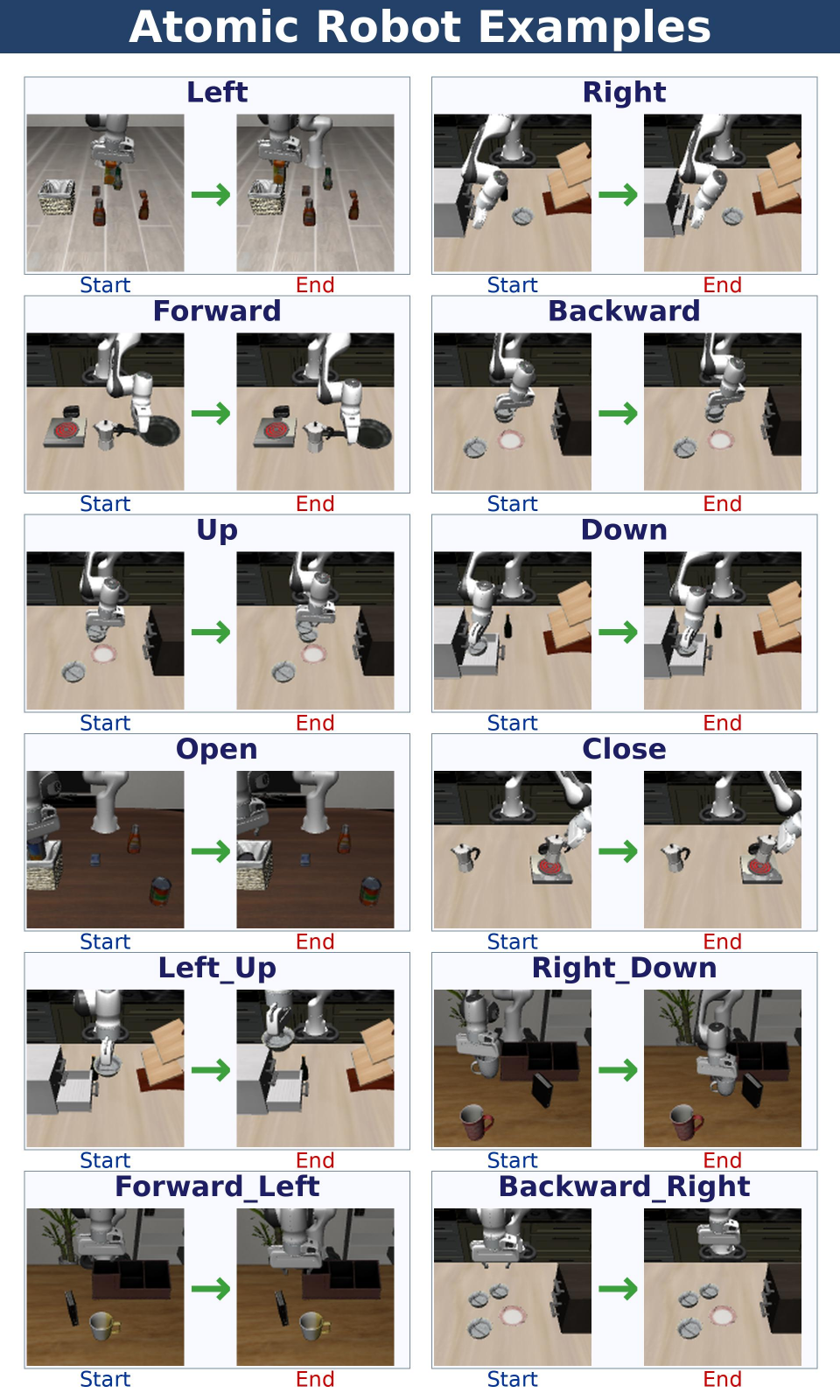}
    \caption{Representative examples of Kinematic Atomic Primitives from the \textit{Atomic Robot} subset. The visualizations demonstrate precise, discrete end-effector movements and gripper state changes in structured exocentric environments.}
    \label{fig:atomic_examples}
\end{figure}

\begin{figure}[htbp]
    \centering
    \includegraphics[width=0.8\linewidth]{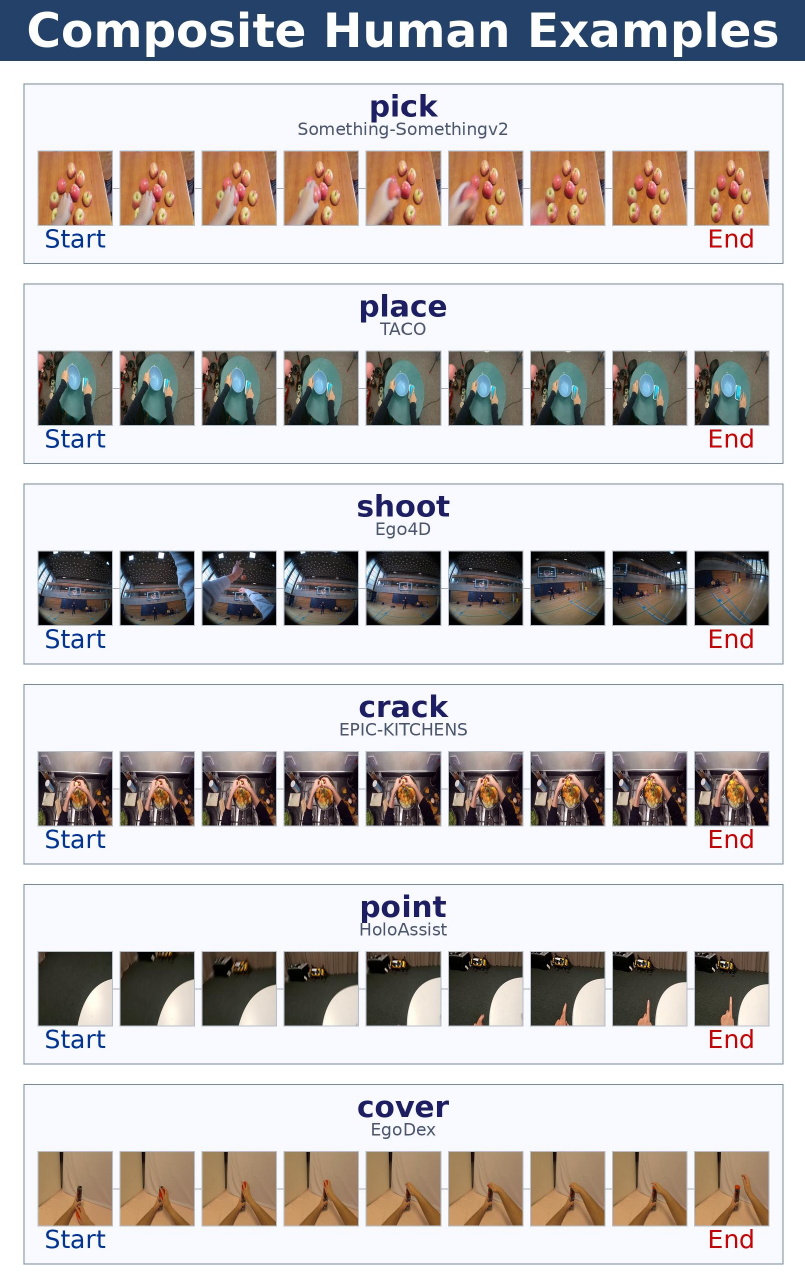}
    \caption{Representative examples of Composite Human Behaviors.}
    \label{fig:composite_human_examples}
\end{figure}

\begin{figure}[htbp]
    \centering
    \includegraphics[width=0.8\linewidth]{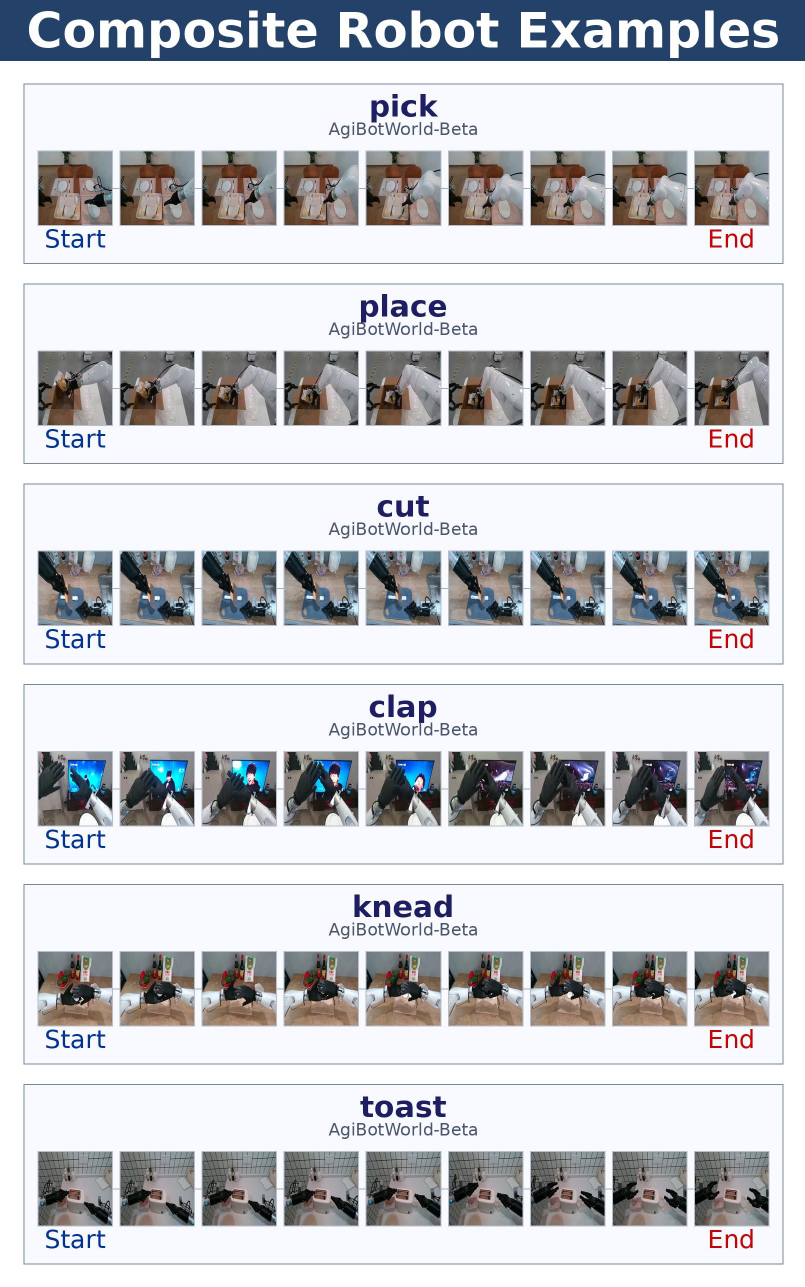}
    \caption{Representative examples of Composite Robot Behaviors sourced from AgiBotWorld-Beta~\citep{agibotworld}.}
    \label{fig:composite_robot_examples}
\end{figure}

This classification track consists of two primary subsets:
\begin{itemize}
    \item \textbf{Kinematic Atomic Primitives (Atomic Robot, 28 classes):} This subset contains 25,940 high-quality exocentric image pairs with trajectories derived from the LIBERO suite. Tasks evaluate 28 discrete kinematic primitives, including directional movements (e.g., \textit{move\_top\_left}, \textit{move\_forward}) and binary end-effector states (\textit{gripper\_open}, \textit{gripper\_close}). Figure~\ref{fig:atomic_examples} illustrates representative examples of these fine-grained spatial displacements and state transitions.
    \item \textbf{Composite Behaviors (Composite Human, 123 classes \& Composite Robot, 54 classes):} This subset evaluates abstract behavioral semantics using 1,190,820 video clips. As detailed in Table~\ref{tab:dataset_composition}, within this composite video subset, the data is inherently balanced across domains, with human videos comprising 54.8\% and robotic manipulations comprising 45.2\%. Representative visual examples encompassing diverse human and robotic embodiments are shown in Figure~\ref{fig:composite_human_examples} and Figure~\ref{fig:composite_robot_examples}.
\end{itemize}

\paragraph{Regression Track}
This track focuses entirely on the continuous low-level execution required for robotic control. As detailed in Table~\ref{tab:dataset_composition}, it comprises 595,237 image pairs and corresponding motion trajectories derived from diverse simulated and real-world robotic environments.

\begin{table}[htbp]
    \centering
    \caption{Detailed composition and statistics of the LARY Benchmark dataset. The dataset is divided into two primary tracks: Classification (including both composite videos and atomic image pairs) and Regression. Proportions are calculated independently for each track.}
    \label{tab:dataset_composition}
    \resizebox{\textwidth}{!}{
    \begin{tabular}{lcccccc}
        \toprule
        \begin{tabular}[c]{@{}l@{}}\textbf{Source} \\ \textbf{Dataset}\end{tabular} & \textbf{Domain} & \textbf{Viewpoint} & \begin{tabular}[c]{@{}c@{}}\textbf{Task} \\ \textbf{Allocation}\end{tabular} & \begin{tabular}[c]{@{}c@{}}\textbf{Data} \\ \textbf{Format}\end{tabular} & \textbf{Count} & \textbf{Proportion} \\
        \midrule
        \multicolumn{7}{c}{\textbf{Classification Track}} \\
        \midrule
        LIBERO~\citep{libero}              & Robot & Exo & Atomic & Image Pairs & 25,940 & 2.1\% \\
        \cmidrule{1-7}
        EgoDex~\citep{egodex}              & Human & Ego & Composite & Video Clips & 502,752 & 41.3\% \\
        SSv2~\citep{ssv2}                  & Human & Ego & Composite & Video Clips & 106,911 & 8.8\% \\
        Ego4D~\citep{ego4dv2}              & Human & Ego & Composite & Video Clips & 22,744 & 1.9\% \\
        HoloAssist~\citep{HoloAssist2023}  & Human & Ego & Composite & Video Clips & 10,825 & 0.9\% \\
        EPIC-KITCHENS~\citep{epickitchens} & Human & Ego & Composite & Video Clips & 5,879 & 0.5\% \\
        TACO~\citep{taco}                  & Human & Ego & Composite & Video Clips & 3,286 & 0.3\% \\
        \cmidrule{1-7}
        AgiBotWorld-Beta~\citep{agibotworld}& Robot & Ego & Composite & Video Clips & 538,423 & 44.2\% \\
        \midrule
        \multicolumn{7}{r}{\textit{Total Classification Samples: 1,216,760}} \\
        \midrule
        \multicolumn{7}{c}{\textbf{Physical Regression Track}} \\
        \midrule
        CALVIN~\citep{calvin}              & Robot & Exo & Regression & Trajectories & 209,921 & 35.3\% \\
        VLABench~\citep{vlabench}          & Robot & Exo & Regression & Trajectories & 112,030 & 18.8\% \\
        RoboCOIN~\citep{robocoin}          & Robot & Ego & Regression & Trajectories & 148,767 & 25.0\% \\
        AgiBotWorld-Beta~\citep{agibotworld}& Robot & Ego & Regression & Trajectories & 124,519 & 20.9\% \\
        \midrule
        \multicolumn{7}{r}{\textit{Total Regression Trajectories: 595,237}} \\
        \bottomrule
    \end{tabular}
    }
\end{table}

\subsection{Additional Data Curation Details}
Existing embodied datasets exhibit vast disparities in temporal boundaries and semantic annotations. To unify these heterogeneous sources, we designed an automated, multi-stage data processing engine.

\paragraph{Atomic Classification}
We identify discrete robotic arm movements within the LIBERO dataset by tracking the cumulative positional offset of the end-effector. We focus primarily on fundamental translations and binary gripper states. Specifically, we extract the precise start and end frames corresponding to the exact moments when the cumulative displacement surpasses a predefined threshold. Ground-truth action labels are subsequently assigned based on the directional shifts along the axes. This deterministic protocol yields high-quality data samples, each comprising a temporally aligned image pair paired with a kinematic action annotation.

\paragraph{Composite Classification}
The data curation pipeline operates in four primary stages. To ensure absolute annotation consistency across the entire dataset, we uniformly employ the doubao-1.5-pro-vision API for all processing steps:

\begin{enumerate}
\item \textbf{Action Segmentation:} Initially, we perform temporal action segmentation on the raw video sequences using the API. This process yields a massive corpus of short video clips, each paired with a brief, sentence-level action annotation describing the ongoing interaction.
\item \textbf{Video-Description Matching:} Given the inevitable noise from automated cropping and preliminary captioning, we implement a rigorous API-driven filtering protocol to sift the initial clips. Clips are retained strictly based on three criteria: (i) \textit{Temporal validity:} Clip durations must fall within the [0.5s, 20s] interval. This discards clips that are too short to capture complete action semantics, or excessively long clips containing multiple entangled actions. (ii) \textit{Semantic alignment:} The visual content must perfectly match the descriptive sentence, ensuring the clip captures a single, complete action rather than fragmented or disjointed sequences. (iii) \textit{Perspective consistency:} We strictly retain only egocentric videos, explicitly filtering out exocentric viewpoints from mixed datasets (e.g., Something-Somethingv2~\citep{ssv2}). Concurrently, the model extracts the core action verb from the descriptive sentence. This stage outputs a refined set of video clips paired with extracted verb labels.

\item \textbf{Video-Verb Consistency Check:} To eliminate semantic ambiguity and ensure that the isolated verb accurately encapsulates the visual content, we conduct a secondary API-based verification. This step explicitly evaluates whether the single verb token alone maintains strict visual-semantic alignment with the corresponding video clip, discarding any poorly correlated pairs.

\item \textbf{Manual Sampling Inspection:} Finally, we perform a manual quality assurance review to exclude verbs that lack explicit or well-defined kinematic meanings. Action categories representing overly abstract or kinematically ambiguous operations (e.g., \textit{apply}, \textit{arrange}, \textit{clean}) are systematically purged from the taxonomy to maintain the physical and dynamic rigor of the benchmark.
\end{enumerate}

We provide the exact prompts used with doubao-1-5-thinking-vision-pro-250428 to trim and check the videos. Sampling FPS is set to 5, and the minimum number of sampling frames is 16 to ensure it can capture the precise duration of the action.
\begin{PromptBox}
\textbf{[System Prompt]}
You will be provided with \{n_frames\} separate frames uniformly sampled from a video with their 
sampled time.

\textbf{[Action Segmentation Prompt]}
<video_1>Please watch this video and find out all key actions in this video. Ensure each action 
is processed independently of the others. List every single action separately with its start/end 
timestamps. Keep descriptions concise, objective, and in English.

\textbf{[Video Description Matching Prompt]}
<video_1>Video Analysis Task (Strict Matching):
1. Strict Verification: Does the video content EXCLUSIVELY represent the description \{old_desc\}?
   - Match if: The video contains the described action and NOTHING else.
   - Mismatch if: The video is missing parts of the action, OR contains any additional, unrelated 
   actions before, during, or after.
2. Identify perspective: '1st' (ego) or '3rd' (non-ego).
3. Final Action: If and only if it's a STRICT match, provide ONE most precise English verb 
defining the movement, not limited to the exact word in the description (e.g., 'take'); 
Otherwise, return 'None'.
Return ONLY JSON: \{\{"perspective": "1st/3rd", "action": "verb/None"\}\}

\textbf{[Video-Verb Consistency Check Prompt]}
<video_1>Please watch this video and determine if the action \{action\} is performed.
Output Yes only if the action is clearly identifiable and the video content does not contradict 
the label. Output No if the action is missing, incorrect, or represents a different verb entirely.
Return ONLY JSON: \{\{"output": "Yes/No"\}\}

\end{PromptBox}

\paragraph{Regression}
We obtain the requisite image pairs and action trajectories by sampling observations at a fixed temporal stride. The start and end frames are paired with the continuous sequence of absolute kinematic actions occurring strictly between them.

\section{Experimental Details of General LAMs}
\label{sec:Experiment_Details}

\subsection{Models and Settings}
\label{sec:models_and_settings}

To rigorously evaluate the impact of different visual priors on latent action learning, we design four variants of General Latent Action Models (General LAMs): LAPA-DINOv2, LAPA-DINOv3, LAPA-SigLIP2, and LAPA-MAGVIT2. Across all variants, the weights of the pre-trained visual backbones are kept frozen. 

For the first three understanding-oriented models (DINOv2~\citep{dinov2}, DINOv3~\citep{dinov3}, and SigLIP2~\citep{siglip2}), we substitute the pixel-level inputs of the original LAPA VQ-VAE architecture with continuous feature embeddings extracted from the penultimate layers of their respective vision encoders. Consequently, their training objective is formulated to reconstruct these high-level latent features rather than raw pixels. Conversely, LAPA-MAGVIT2 operates as a generative-based variant, where the reconstruction target remains in the visual pixel space.

\subsection{Training Configurations and Hyperparameters}
\label{sec:training_configurations}

All General LAM variants are trained on an expansive internal video dataset comprising over 2 million frames. This corpus is meticulously curated to encompass a diverse mixture of human demonstrations, robotic manipulations, and general environmental dynamics. During the training phase, we extract the start and end frames by sampling with a randomized temporal stride within a predefined range. This stochastic sampling strategy prevents overfitting to specific frame rates and forces the latent representations to capture robust temporal dynamics across varying motion speeds. All training workloads were executed utilizing 8 NVIDIA H800 (141GB) GPUs.

The input image resolution is standardized to $224 \times 224$. The latent action quantization modules consistently employ a symmetric spatial-temporal transformer architecture. Both the spatial and temporal encoders are configured with a depth of 4 layers, utilizing 16 attention heads with a dimension of 64 per head. All General LAM variants are trained for exactly 100,000 steps and share a foundational quantization configuration, which includes a quantization dimension of 32, a codebook size of 8, and a sequence length of 16. Regarding the encoder-specific spatial parameters, both LAPA-DINOv3 and LAPA-DINOv2 process features with an input dimension of 1024; however, they utilize patch sizes of 16 and 14, respectively. LAPA-SigLIP2 operates on an input feature dimension of 768 with a patch size of 16. LAPA-MAGVIT2 is consistently configured with a patch size of 16, processing an input feature dimension of 18.

\section{Additional Visualizations and Case Studies}
\label{sec:Additional_Visualizations}

\subsection{Cross-Embodiment Gap Across Different Model Paradigms}
\label{sec:cross_embodiment_gap}

To further investigate the morphological domain gap between human and robotic embodiments, we analyze how different representation paradigms handle shared semantic actions (e.g., \textit{pick}, \textit{place}, \textit{push}, \textit{pull}) present in both the \textit{Composite Human} and \textit{Composite Robot} subsets. Although these actions share identical high-level intentions, their visual appearances diverge significantly due to differences in end-effector morphologies.

\begin{figure}[htbp]
    \centering
    \includegraphics[width=\linewidth]{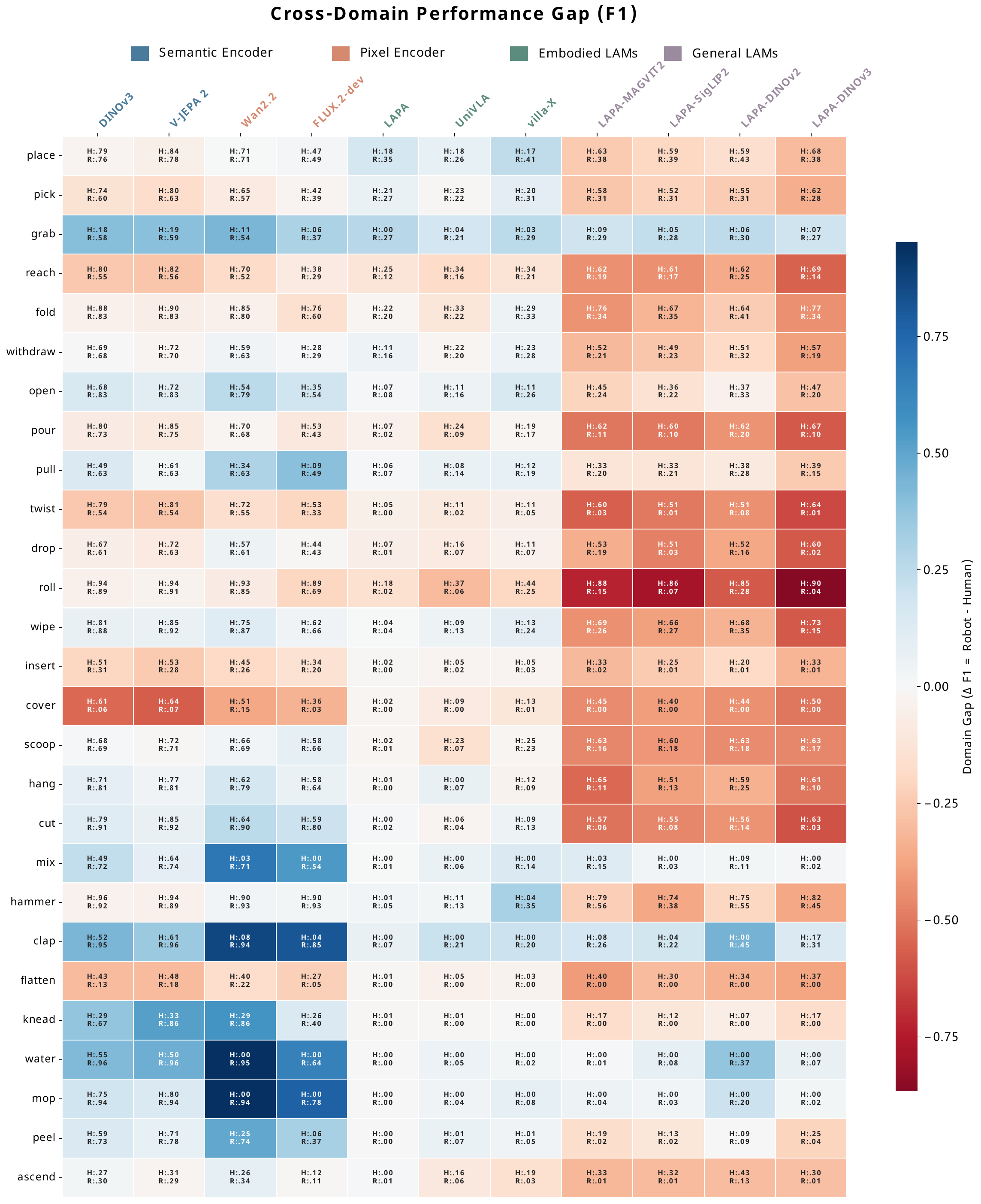}
    \caption{
        \textbf{Cross-domain performance gap analysis across shared action categories.} 
        The heatmap visualizes the F1 score discrepancy ($\Delta \text{F1} = \text{Robot} - \text{Human}$) between human and robot domains. Red regions highlight superior performance on human actions, whereas blue regions indicate better performance on the corresponding robot actions. Absolute F1 scores for Human (H) and Robot (R) are explicitly reported within each cell.
    }
    \label{fig:cross_embodiment_models}
\end{figure}
As illustrated in Figure~\ref{fig:cross_embodiment_models}, the cross-domain gaps reveal four critical insights:

\begin{itemize}
    \item \textbf{General LAMs Exhibit a Human-Centric Preference:} While our General LAMs (e.g., LAPA-DINOv3, LAPA-SigLIP2) successfully elevate the overall F1 bounds compared to standard Embodied LAMs, they demonstrate a distinct preference for human embodiments. For the majority of actions, such as \textit{twist}, \textit{drop}, and \textit{roll}, the human F1 scores significantly outstrip their robotic counterparts (e.g., for \textit{roll}, LAPA-DINOv3 achieves H: 0.90 vs. R: 0.04). This phenomenon suggests the General LAMs effectively learn and capture a much richer set of human behaviors, likely benefiting from the extensive human-centric priors embedded within the frozen visual backbones.
    
    \item \textbf{Understanding Encoders Demonstrate Balanced Cross-Domain Robustness:} In stark contrast to the LAMs, continuous foundational vision encoders, particularly DINOv3, emerge as the most morphologically balanced paradigms. DINOv3 maintains highly comparable and competitive F1 scores across both domains for a wide array of categories (e.g., \textit{place}, \textit{fold}, \textit{scoop}). This demonstrates that robust visual priors possess a powerful, domain-agnostic capability to understand interactions, irrespective of the specific end-effector morphology.
    
    \item \textbf{Embodied LAMs Show a Marginal Edge on Robotic Domains:} While specialized Embodied LAMs (e.g., LAPA, UniVLA) generally struggle with low overall precision across both domains, they exhibit a slight performance edge on robotic actions compared to human ones. This marginal robotic preference is fundamentally because these architectures are primarily trained on robotic manipulation datasets, thereby lacking the diverse human-centric exposure required for broader cross-domain generalization.
    
    \item \textbf{Data Imbalance Drives Robotic Bias in Specific Actions:} Interestingly, a few specific actions, namely \textit{grab} and \textit{mix}, consistently show a distinct bias towards the robotic domain across all models. This anomalous robotic preference is primarily attributed to the underlying dataset distribution. As previously illustrated in Figure~\ref{fig:action_distribution}), these specific interactions are substantially overrepresented in the robotic training sets compared to the human datasets, structurally skewing the learned representations even for general vision models.
\end{itemize}

\subsection{Spatiotemporal Grounding via Attention Visualization}
\label{sec:supp_attention_maps}

Complementing the analysis in Section 5.2 of the main text, we provide an extended gallery of cross-attention visualizations.

\begin{figure}[htbp]
    \centering
    \includegraphics[width=\textwidth]{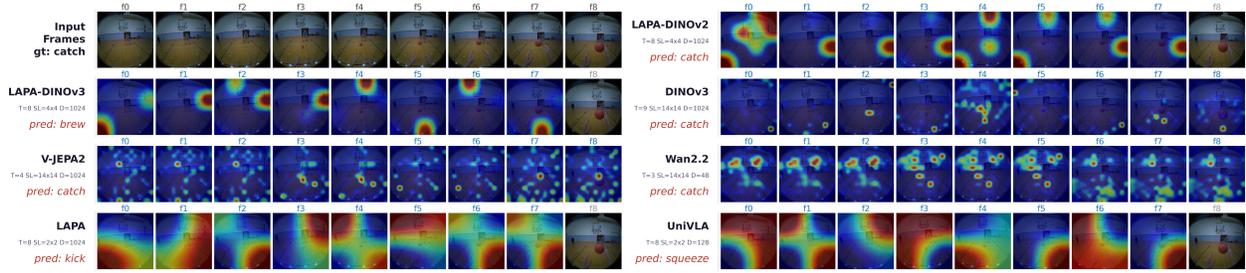}
    \caption{A case for the action \textit{catch}.}
    \label{fig:attn_catch}
\end{figure}


\begin{figure}[htbp]
    \centering
    \includegraphics[width=\textwidth]{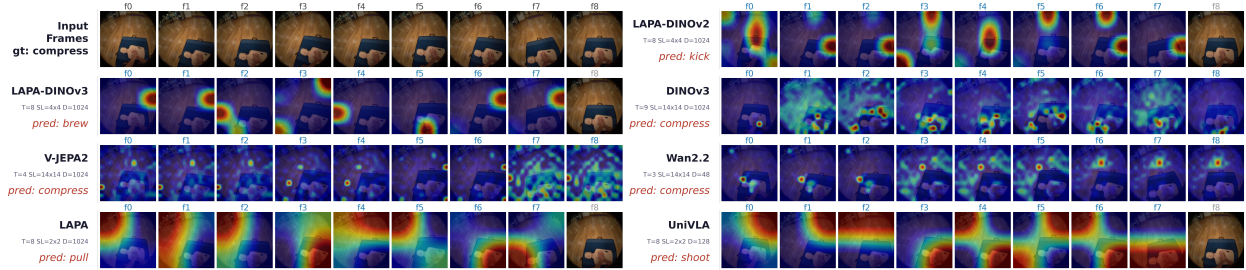}
    \caption{A case for the action \textit{compress}.}
    \label{fig:attn_compress}
\end{figure}

\begin{figure}[htbp]
    \centering
    \includegraphics[width=\textwidth]{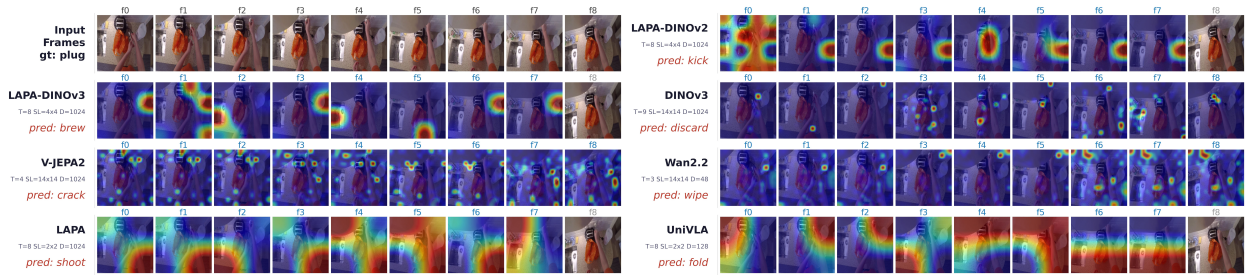}
    \caption{A case for the action \textit{plug}.All models failed.}
    \label{fig:attn_plug}
\end{figure}

\begin{figure}[htbp]
    \centering
    \includegraphics[width=\textwidth]{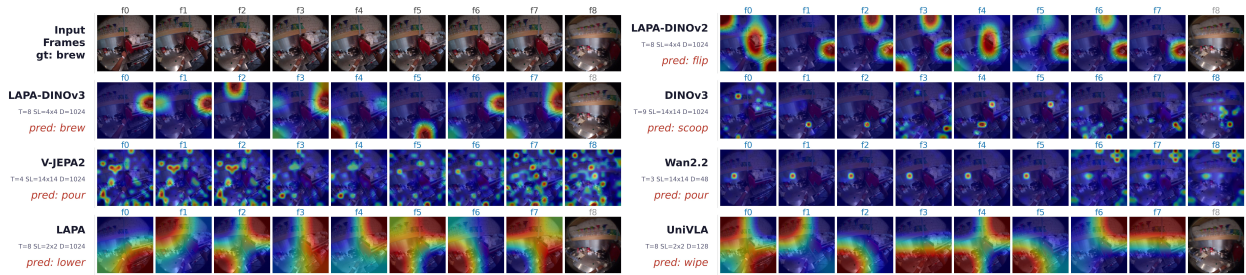}
    \caption{A case for the action \textit{brew}.}
    \label{fig:attn_brew}
\end{figure}

\end{document}